\documentclass{article}
\usepackage{log_2025}						

\usepackage{duckuments}						

\usepackage[utf8]{inputenc} 
\usepackage[T1]{fontenc}    
\usepackage{url}            
\usepackage{booktabs}       
\usepackage{amsfonts}       
\usepackage{nicefrac}       
\usepackage{microtype}      
\usepackage{xcolor} 
\usepackage{subcaption}
\usepackage{wrapfig}
\usepackage{multirow}
\usepackage{placeins}
\usepackage{graphicx} 
\usepackage{amsmath}
\usepackage{amssymb}
\usepackage[dvipsnames]{xcolor}

\usepackage{Definitions}

\usepackage[numbers,compress,sort]{natbib}	

\title{Integrating Sequential and Relational Modeling for User Events: Datasets and Prediction Tasks}

\author[Fathony et al.]{%
  Rizal Fathony, Igor Melnyk, Owen Reinert, Nam H. Nguyen, Daniele Rosa, C. Bayan Bruss \\
  Capital One \\
  \texttt{\{rizal.fathony,igor.melnyk,owen.reinert,nam.nguyen,} \\ \texttt{daniele.rosa,bayan.bruss\}@capitalone.com} 
}

\begin{document}

\maketitle

\begin{abstract}
\looseness=-1
User event modeling plays a central role in many machine learning applications, with use cases spanning e-commerce, social media, finance, cybersecurity, and other domains. User events can be broadly categorized into personal events, which involve individual actions, and relational events, which involve interactions between two users. These two types of events are typically modeled separately, using sequence-based methods for personal events and graph-based methods for relational events. Despite the need to capture both event types in real-world systems, prior work has rarely considered them together. This is often due to the convenient simplification that user behavior can be adequately represented by a single formalization, either as a sequence or a graph. To address this gap, there is a need for public datasets and prediction tasks that explicitly incorporate both personal and relational events. In this work, we introduce a collection of such datasets, propose a unified formalization, and empirically show that models benefit from incorporating both event types. Our results also indicate that current methods leave notable room for improvements. We release these resources\footnote{\href{ https://huggingface.co/datasets/capitalone/PRES}{\underline{Link to the datasets and prediction tasks}}. \href{https://github.com/CapitalOne-Research/personal-relational-event-sequence}{\underline{Link to the dataset construction and experimentation code}.}
} to support further research in unified user event modeling and encourage progress in this direction.
\end{abstract}


\section{Introduction}





\looseness=-1
Modeling user events is a central task in machine learning with broad applications across various domains 
\citep{purificato2024user, martin2021survey,deng2024advances}.
In e-commerce, it is used to capture user preferences for personalized ranking and product recommendation 
\citep{he2023survey, zeng2019user}. 
In social media platforms, event modeling supports feed optimization and engagement prediction by inferring user interests over time 
\citep{abdelhafez2013survey, piao2018inferring, fang2023snapchat}. 
Financial systems leverage user behavior data for fraud detection, credit risk assessment, and behavioral profiling 
\citep{hernandez2024fraud, mashrur2020risk, ozbayoglu2020deep, hojaji2022behavioral}. 
Online services such as search and streaming platforms rely on user event sequences for content recommendation under real-time constraints 
\citep{zhao2023mbsrs, gayoavello2009session, covington2016youtube, amatriain2013netflix}.
In cybersecurity, modeling user and system events is essential for detecting anomalies and preventing intrusions \citep{lashkari2017survey, wang2017clickstream}. These applications demonstrate the importance of building models that can effectively capture complex, context-dependent user behavior from event sequences.

User events can be broadly categorized into personal and relational events. Personal events involve only a single user and reflect individual actions, such as searching for content, viewing items, or posting updates. In contrast, relational events involve interactions between two or more users, such as following another user, co-editing a document, or exchanging messages. Traditionally, these two types of events are often modeled separately. Relational events are commonly modeled using graph-based approaches that capture structural dependencies and interaction patterns among users 
\citep{gao2023survey,guo2020survey,wu2022graph,li2024graph}. 
On the other hand, personal events are typically modeled as sequences using recurrent or attention-based architectures to capture temporal dependencies in personal event histories 
\citep{Chen2024ASO,pan2024survey,boka2024survey,sun2019bert4rec,Tan2016ImprovedRN,Kang2018SelfAttentiveSR,liguori2023modeling}.

\begin{figure}[t]
 \centering
  \includegraphics[width=0.85\textwidth]{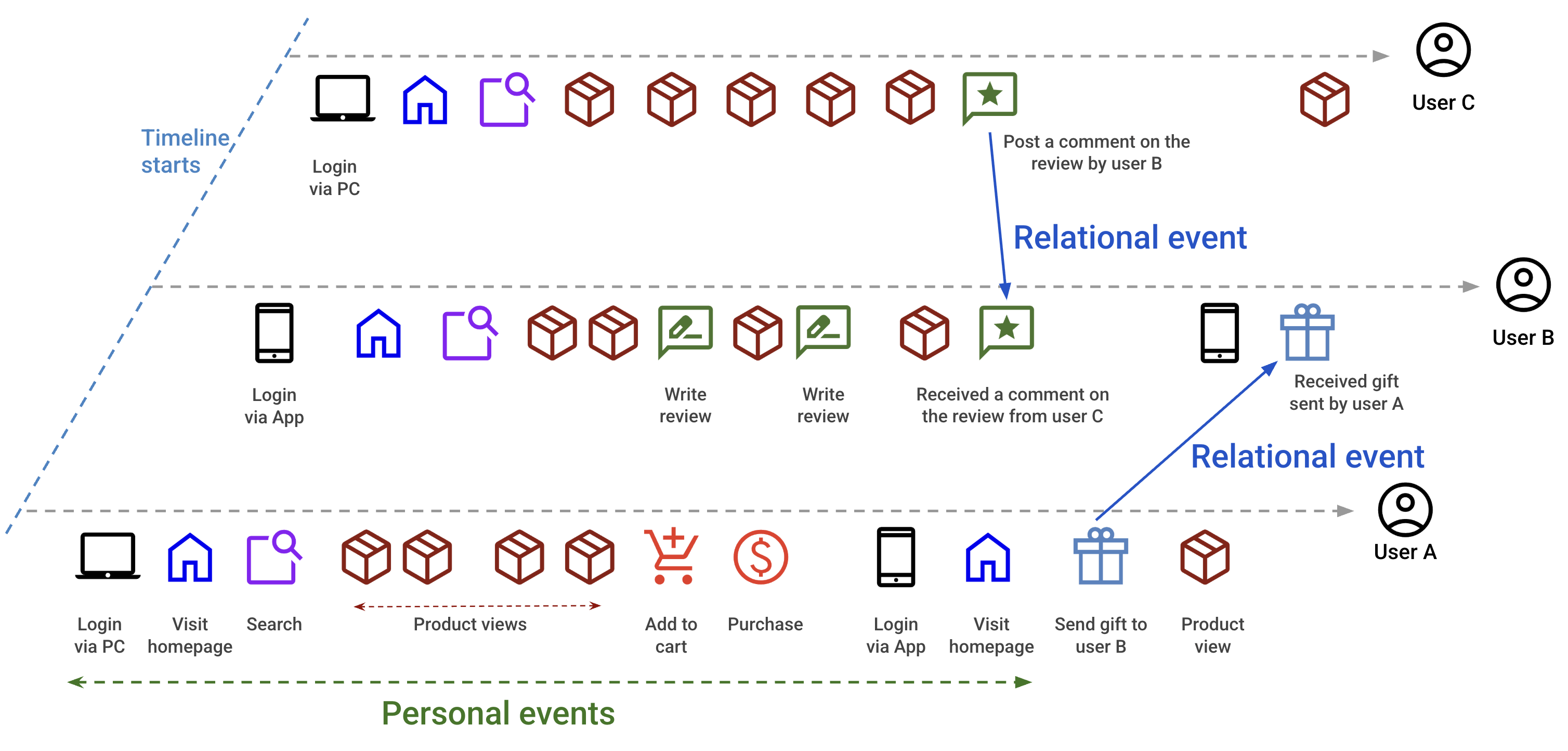}
  \caption{An illustration of personal and relational events in e-commerce. Personal events involve a single user, such as login, search, view, or purchase. Relational events involve interaction between two users, such as sending a gift or commenting on another user's review.}
  \label{fig:pres}
\end{figure}

There have been efforts in the graph area to capture both structural and temporal dependencies using temporal graph formalizations (such as CTDG \citep{nguyen2018continuous,tgn_icml_grl2020,kazemi2020representation}) and models built on top of these formalizations (such as TGAT \citep{xu2020inductive}, TGN \citep{tgn_icml_grl2020},  DyRep \citep{trivedi2019dyrep}, TNCN \cite{zhang2406efficient}, and others  \cite{luo2022neighborhood,yu2023towards,gravina2024long}). However, these approaches primarily focus on the temporal dependencies of relational events while neglecting personal events. For example, the formalization used in the Temporal Graph Benchmark (TGB) papers \citep{huang2023temporal,gastinger2024tgb} and recent temporal graph models \citep{zhang2406efficient,luo2022neighborhood,yu2023towards,gravina2024long}
defines a temporal graph as a stream of triplets consisting of source, destination, and timestamp. Personal events that involve only a single entity cannot be directly represented under this formulation. One workaround is to convert all personal events into nodes and define personal events as triplets of user node, event node, and timestamp. However, this construction is not as straightforward for capturing temporal dependencies in personal event histories compared to sequence-based modeling.

Going back to the personal and relational event category, in many application domains, the number of personal events is typically much larger than that of relational events. 
For example, 
in e-commerce platforms, as illustrated in Figure \ref{fig:pres}, users often view products, search for items, or add products to their cart, whereas relational interactions, such as referrals, sending gifts, or socially engaged reviews, are less frequent. 
In financial systems, customers routinely perform account queries, check balances, or initiate transactions, while peer-to-peer interactions such as money transfers or joint account actions are relatively infrequent. 
In cybersecurity systems, personal events may include actions like logging in, accessing files, or executing processes, while relational events, such as remote connections to other users, or file sharing between users, occur less frequently. 
Despite their higher volume, personal events are often underrepresented in existing graph-based formulations, which tend to prioritize relational structure. In practice, however, both personal and relational events carry complementary signals, and many predictive tasks, such as item recommendation, fraud detection, customer profiling, and behavior forecasting, benefit from capturing both types of information.

\looseness=-1
Even though there is a need to capture both personal and relational events in many application domains, prior work has rarely considered them together. Practitioners often simplify the complexity of user event modeling by adopting either a graph or a sequence formalization, as most machine learning models are developed within one of these frameworks. As a result, one type of event, typically the less convenient to represent, is often ignored entirely, leading to an incomplete view of user behavior. To build a more comprehensive understanding of user event modeling, there is a need for public datasets and benchmark tasks that explicitly incorporate both event types. Such resources would provide a foundation for developing and evaluating models that integrate these complementary signals.

\looseness=-1
\paragraph{Summary of Contributions.} 
In this work, we aim to support the study of user event modeling that incorporates both personal and relational events. Our contributions are as follows:
\begin{itemize}
\setlength\itemsep{0pt}
\setlength\topsep{0pt}
    \item We curate, pre-process, and release a collection of public datasets and prediction tasks that explicitly include both personal and relational events.
    \item We introduce a new formalization for user event modeling that captures both personal and relational events.
    \item We empirically demonstrate that incorporating both personal and relational events improves performance on a range of prediction tasks.
    \item We show that existing models, originally developed for either sequential or relational data only, are less well suited for this event modeling setting, leaving room for future improvements.
    \item We invite the community to use these resources and close the gap in unified user event modeling.
\end{itemize}

\section{Related Works}

\label{sec:related_works}

We discuss the most relevant works in this section, while a more comprehensive list of the related works can be seen in Appendix \ref{sec:add_rel_work}.


\textbf{Event sequence} modeling, such as temporal point processes \citep{mei2017neural,zuo2020transformer,zhang2020self}, deals with event stream data $(e_1, t_1), (e_2, t_2), \cdots$, where each $e_i$ is an event type drawn from an event set $\Ecal = \{1,2,\cdots,|\Ecal|\}$, and $0 \leq t_1 \leq t_2 \leq \cdots$ are the times of occurrence. 
In \textbf{user event sequence} modeling, such as personalized event prediction tasks \citep{letham2013sequential,boyd2020user,yang2022towards}, we have $N$ users, denoted as $\Ucal = \{u_1, u_2, \cdots, u_N\}$. Each user has their own sequence of events, denoted $\text{Seq}(u) = [(e_1,t_1)^{(u)}, (e_2,t_2)^{(u)}, \cdots]$. In some applications, exact timestamps are replaced with discrete time steps, simplifying the definition to $\text{Seq}(u) = [e_1^{(u)}, e_2^{(u)}, \cdots]$. 
In other variants of sequence modeling, the event space may be restricted to a single domain, for example, products consumed by a user in sequential recommendation tasks \citep{hidasi2015session,kang2018self,sun2019bert4rec}. 
While these user event sequence models capture the complexity of event sequences within a user, they lack the ability to encode user-to-user interactions.

\looseness=-1
\textbf{Graph modeling}, on the other hand, explicitly encodes user-to-user interactions through its node and edge abstraction, $\Gcal = ( \Vcal, \Ecal )$, where $\Vcal$ and $\Ecal$ denote the node and edge sets, respectively. Temporal graph abstractions, such as CTDG \citep{nguyen2018continuous,tgn_icml_grl2020}, incorporate temporal dynamics by representing a graph as a sequence of time-stamped events $\Gcal = \{x(t_1), x(t_2), \cdots\}$. Each event $x(t)$ can be either (1) a \textit{node-wise event} (e.g., node addition or feature update) represented by a feature vector $\vvec_i(t)$, where $i$ is the node index, or (2) an \textit{interaction event} between node $i$ and node $j$ represented by a temporal edge $\evec_{i,j}(t)$. 
Another CTDG formulation by \citet{kazemi2020representation} describes a temporal graph as a pair consisting of an initial graph and an observation set $(\Gcal, \Ocal)$. The observation set describes the evolution of the graph, with each observation is represented as a tuple (\textit{event type, event, timestamp}), where the event type may include edge addition, edge deletion, node addition, node deletion, node splitting, node merging, etc. 
In practice, however, many temporal graph models \citep{zhang2406efficient,luo2022neighborhood,yu2023towards,gravina2024long} omit node-level events altogether and instead define a temporal graph as a sequence of interaction events expressed as triplets, $\Gcal = \{(u_1,v_1,t_1), (u_2,v_2,t_2), \cdots\}$, where $u$ and $v$ represent the source and destination nodes.

\section{Problem Formalization}
\label{sec:formulation}


\paragraph{Notations.}

\looseness=-1
For our \textbf{P}ersonal and \textbf{R}elational User \textbf{E}vent \textbf{S}equence (PRES) modeling, 
we take inspiration from the standard user event sequence modeling. 
We denote a set of users (which can also be a customer, account, entity, etc.) as $\Ucal = \{ u_1, u_2, \cdots, u_{N} \}$, where $N$ is the number of users. Each user has their own sequence of events that occur over time. For example, the sequence for user $u_i$ is denoted as $\text{Seq}(u_i) = [(e_1,t_1), (e_2,t_2), \cdots, (e_{M_i}, t_{M_i})] $, where $e$ describes an event, $t$ describes the time at which the event occurs, and $M_i$ denotes the number of events for user $u_i$. Each user may have a different number of events in their event sequence. We denote the set of all user sequences by $\Scal = \left\{\text{Seq}(u) \mid u \in \Ucal\right\}$.
An event may come from two different event sets: the \textit{personal event} set and the \textit{relational event} set. The personal event set contains a set of events that can occur for an individual user; $p \in \Pcal \triangleq \{1,2,\cdots,|\Pcal|\}$. 
The relational event set contains a set of all possible events $r \in \Rcal \triangleq \{1,2,\cdots,|\Rcal|\}$, which involve a relation from one user to another. Thus, an event can be defined by a personal event $e = p$, or a relational event tuple $e = (r, v)$, where $v$ is another user.



\paragraph{Difference from temporal graph formulation.} Although motivated by a similar goal of integrating temporal and relational information, our formulation differs from temporal graphs in several ways:
\begin{itemize}
\setlength\itemsep{0.3pt}
\setlength\topsep{0.3pt}
    \looseness=-1
    \item In the temporal graph formulation, events are ordered globally across all users, similar to the standard (non-user-based) event sequence modeling. In contrast, PRES follows personalized user-event sequence modeling, where each user has their own separate sequence of events.
    
    \item The temporal graph formulation places greater emphasis on modeling interaction events between two nodes, often neglecting user events that do not involve another user. Many recent temporal graph models, such as TNCN \citep{zhang2406efficient}, NAT \cite{luo2022neighborhood}, DyGFormer \citep{yu2023towards}, and others \citep{gravina2024long,lu2024improving,gao2025hyperevent}; recent temporal graph benchmarks, such as TGB \cite{huang2023temporal} and TGB 2.0 \cite{gastinger2024tgb}; as well as popular graph frameworks such as PyTorch Geometric \citep{fey2019fast}, even reduce the formulation to a sequence of triplets representing interaction events with no support for encoding personal events. By contrast, PRES treats personal events as first-class components. This is particularly relevant in many areas mentioned in the introduction, where the number of personal events is far larger than that of relational events.

    \item Even for the temporal graph formulation that accommodate node-wise events, it does so by updating node feature vectors. This differs from event sequence modeling (including PRES), where each event is drawn from a discrete set of events, a property that may not be easily encoded as feature changes. 
    To represent discrete personal events, temporal graph models may convert the events into nodes and then build a heterogeneous dynamic graph that encodes the event as an interaction between a user node and this newly created event node. In some events, such as ‘viewing product’, this approach works fine. However, it does not work well for representing personal events that describe intransitive actions (e.g., ‘sign in’, ‘login’, or ‘subscribe’) or status changes (e.g., ‘payment successful’, ‘login unsuccessful’, or 'request denied').
    
    \item In graph models, every entity must be a node. This becomes problematic when the entity has a hierarchical structure. For example, in a location-based social network application, when modeling a user’s location check-in event, one may want to capture hierarchical information such as continent, country, state, city, neighborhood, and block. Representing the location check-in as a single node in a graph model loses this hierarchical structure. In contrast, PRES allows more flexibility: using a sequence model, events can be freely tokenized. A check-in event can be split into multiple tokens, each corresponding to a different level of location granularity.
\end{itemize}
When we compare with the CTDG formulation by \citet{kazemi2020representation}, all the concerns above remain. While this formulation allows more flexible observations, it is still restricted to events that describe the evolution of the graph through changes in nodes or edges, such as edge additions, edge deletions, node additions, node deletions, node splitting, or node merging.

\section{Datasets and Prediction Tasks}

\label{sec:dataset}

\subsection{Dataset Information}


\begin{table}[t]
\centering
\caption{Dataset Statistics}
\label{tbl:data-stats}
{
\footnotesize
\resizebox{\textwidth}{!}{%
\begin{tabular}{@{}l|rrrrr@{}}

\toprule
\textbf{Properties} &  \textsc{brightkite} & \textsc{gowalla} & \textsc{amazon-clothing} & \textsc{amazon-electronics} & \textsc{github} \\
\midrule
Personal Events       & check-in    & check-in    & product rating     & product rating     & github activity \\
Relational Events     & friendship          & friendship           & co-review          & co-review          & collaboration \\
\midrule
\# Users              & 58,228              & 196,591              & 185,986            & 254,064            & 3,669,079 \\
\# Events             & 5,130,866           & 8,342,943            & 1,591,947          & 2,938,178          & 102,878,895 \\
\# Personal Events    & 4,702,710           & 6,442,289            & 1,573,869          & 2,281,128          & 95,974,149 \\
\# Relational Events  & 428,156             & 1,900,654            & 18,078             & 657,050            & 6,904,746 \\
\midrule
\# Unique Events      & 628,519             & 1,169,154            & 846,052            & 529,198            & 24 \\
\# Unique Timestamps  & 4,506,822           & 5,561,957            & 3,464              & 5,373              & 2,675,990 \\
\midrule
\# Users w. pers. events  & 51,406           & 107,092            & 185,986              & 254,064              & 3,669,079 \\
\# Users w. rel. events  & 58,228           & 196,591            & 5,017              & 49,852              & 441,958 \\
\# Users w. both events  & 51,406           & 107,092            & 5,017              & 49,852              & 441,958 \\
\bottomrule
\end{tabular}
}
}
\end{table}

\looseness=-1
We curated user event datasets from multiple domains\footnote{The datasets presented in this paper are used solely to demonstrate the claims made in this work and are not reflective of any datasets leveraged by Capital One.},
following recent graph dataset curation works \cite{huang2023temporal,gastinger2024tgb,yi2025tgb,hu2020open,hu2021ogb}, and then processed them according to our formalization in Section~\ref{sec:formulation}. The data is stored in CSV format with the columns: \texttt{uid}, \texttt{timestamp}, \texttt{event\_set}, \texttt{event}, and \texttt{other\_uid} (See Appendix \ref{sec:ap-dataset-stats} for details). The \texttt{uid} is a numerical user ID, whereas \texttt{event\_set} indicates whether the event is \textit{personal} or \textit{relational}. For relational events, \texttt{other\_uid} refers to the other user involved in the relation; for personal events, this column is null.
The statistics of each datasets is shown in Table \ref{tbl:data-stats}.
More details on collection and dataset license are available in Appendix \ref{sec:ap-dataset}. The data   processing code are open-sourced and descriptions are available in Appendix \ref{sec:ap-dataset-construction}.


\underline{\textsc{pres-brightkite}}. This dataset contains location check-ins and friendship history of Brightkite users, a location-based social networking platform. It was originally collected by \citet{cho2011friendship} and published in the \href{https://snap.stanford.edu/data/loc-brightkite.html}{SNAP Dataset Repository} \cite{snapnets}. Personal events consist of sequences of location check-ins. We convert the original latitude and longitude coordinates into Geohash-8 representations \cite{niemeyer2008geohash, morton1966computer}, short alphanumeric strings encoding geographic locations. Nearby locations share similar geohash prefixes, while distant ones differ. Example geohashes include {9v6kpmr1}, {gcpwkeq6}, and {u0yhxgm1}. Relational events capture friendship connections among users. The dataset includes 58,228 users and 5,130,866 events. 
In this dataset, only the personal events include timestamp information; relational events do not have associated timestamps.

\looseness=-1
\underline{\textsc{pres-gowalla}}. The dataset also contains the location check-in and friendship history of another social network platform, Gowalla. It was also originally collected by \citet{cho2011friendship} and published in the \href{https://snap.stanford.edu/data/loc-Gowalla.html}{SNAP Repository} \cite{snapnets}. We processed and formatted the data following the same approach used for \textsc{pres-brightkite}. The dataset contains personal events from geohash check-ins and relational events from friendship connections, totaling 8,342,943 events from 196,591 users.

\underline{\textsc{pres-amazon-clothing}}. The dataset contains Amazon product reviews and ratings in the \textit{Clothing, Shoes and Jewelry} category, spanning from May 1996 to July 2014. The raw data was originally collected by \citet{mcauley2015image}. In this dataset, we define personal events as sequences of product IDs and ratings reviewed by a user, for example: {B000MLDCZ2:5} and {B001OE3F08:3}. Relational events represent co-review patterns, where two users have co-reviewed at least three products. The dataset contains event sequences from 185,986 users, with a total of 1,591,947 events.

\underline{\textsc{pres-amazon-electronics}}. The dataset contains Amazon product reviews and ratings in the \textit{Electronics} category, originally collected by \citet{mcauley2015image}. As in \textsc{pres-amazon-clothing}, personal events are defined as sequences of product IDs and ratings, while relational events capture co-review patterns. In total, the dataset contains 2,938,178 events from 254,064 users.


\underline{\textsc{pres-github}}. This dataset contains GitHub user activity from January 2025, collected from the \href{https://www.gharchive.org/}{GH Archive}. Personal events include actions such as {Push}, {CreateBranch}, {CreateRepository}, {PullRequestOpened}, {IssuesOpened}, and {Fork}. Relational events represent project collaboration, where two users are linked if both contributed at least five commits or pull requests to the same repository. The dataset includes 102,878,895 events from 3,669,079 users. In this dataset, only personal events include timestamp information; relational events do not have associated timestamps, similar to \textsc{pres-brightkite} and \textsc{pres-gowalla}.

\paragraph{Variability of the datasets.}
As shown in Table~\ref{tbl:data-stats}, the PRES datasets vary significantly across multiple aspects. The number of users ranges from around 58 thousand in \textsc{pres-brightkite} to more than 3.5 million in \textsc{pres-github}. The number of events also varies, from approximately 1.5 million in \textsc{pres-brightkite} to over 100 million in \textsc{pres-github}. The ratio between relational and personal events ranges from around 1:3 in \textsc{pres-gowalla} to approximately 1:80 in \textsc{pres-amazon-clothing}. The number of unique events also differs widely, from just 24 in \textsc{pres-github} to more than 1 million in \textsc{pres-gowalla}. 
In addition, we observe variability in the number of users having personal events, relational events, and both. Some datasets have more users with relational events than with personal events (e.g., \textsc{pres-brightkite}, \textsc{pres-gowalla}), while others show the opposite trend (e.g., \textsc{pres-amazon-clothing}, \textsc{pres-amazon-electronics}, \textsc{pres-github}). These differences in dataset properties present distinct challenges for modeling user events in each dataset.

\subsection{Prediction Tasks}


\looseness=-1
From the PRES datasets, we define two prediction tasks: one for relational events and one for personal events. These tasks are designed to enable fair comparisons between graph-based, sequence-based, and hybrid models. Relational event prediction focuses on predicting future or held-out subset of user-to-user interactions, similar to link prediction. Personal event prediction aims to predict the likelihood of future occurrence of personal events without requiring exact timestamps, for example, predicting the next 20 personal events given a user's first 100. In both tasks, observed events are compared against negative samples drawn from events not associated with the user. For reproducibility, pre-generated negative samples for validation and test sets are provided in the dataset repository.

\paragraph{Relational event prediction tasks.}
The corresponding tasks for \textsc{pres-brightkite} and \textsc{pres-gowalla} involve friend recommendation. We construct the training data by randomly splitting all relational events into 70\% training, 10\% validation, and 20\% test sets. We also generate negative samples for the validation and test sets. Following \citet{gastinger2024tgb}, we adopt a \textit{1-vs-1000} negative sampling scheme, in which 1,000 negative events are sampled for each relational event in the prediction set. Negative samples are drawn via uniform random sampling of users, excluding those who already have relational events with the target user in the training set.

For the \textsc{pres-github} dataset, the relational event prediction task is defined as collaboration prediction, which involves predicting which users collaborate with a given user. The train, validation, and test splits follow the same procedure as in \textsc{pres-brightkite}, including the sampling method. However, due to the large size of the dataset, we adopt a \textit{1-vs-300} negative sampling scheme.


\looseness=-1
For the \textsc{pres-amazon-clothing} and \textsc{pres-amazon-electronics} datasets, the task is predicting co-review relationships, i.e., which users share at least three products they reviewed. Co-review patterns can reveal how one account may be related to another, which in some cases can help detect fraudulent review syndicates. In these datasets, relational events have timestamp information, i.e., the first time the co-review condition is met. As such, the train, validation, and test splits respect event timestamps. Specifically, we split each user's relational events by taking the last 20\% for test, the previous 10\% for validation, and the rest for training. To manage large histories of some users, we cap test and val sets at 20 and 10 events per user, respectively. Personal events are also split into `observed' and `unobserved' sets based on the timestamp cut-off in the relational event split, with only the observed set used for training. As in \textsc{pres-brightkite}, we adopt a \textit{1-vs-1000} negative sampling scheme.


\looseness=-1
\paragraph{Personal event prediction tasks.}
The task for \textsc{pres-brightkite} and \textsc{pres-gowalla} is to predict the likelihood of a user checking in at a given geohash location in the future. 
We split each user's personal events by taking the last 20\% for test, the previous 10\% for validation, and the rest for training. We also cap the number of events in the test and validation sets to at most 20 and 10 per user, respectively. 
Since personal events are more frequent than relational ones, we adopt a \textit{1-vs-500} negative sampling scheme. As geohash strings encode hierarchical spatial information (e.g., earlier characters represent broader regions), we apply stratified hierarchical sampling. Specifically, negatives are stratified by shared geohash prefixes, from matching the first five characters to none, ensuring a mix of nearby and distant locations.


For the 
\textsc{pres-amazon-clothing} and \textsc{pres-amazon-electronics} 
datasets, the task is to predict future products a user will review and the corresponding ratings, as denoted in their personal event data. We adopt the same train/val/test split strategy as in \textsc{pres-brightkite}, along with a \textit{1-vs-500} negative sampling scheme. Negative samples for each personal event (e.g., {B001OE3F08:3}) are drawn from three sources: (1) the same product with different ratings (e.g., 
{B001OE3F08:1}, 
{B001OE3F08:5}); (2) other personal events not in the user’s training data; and (3) samples from the second set with randomly perturbed ratings.

In the \textsc{pres-github} dataset, the number of unique events in the personal event set is only 24, corresponding to the list of possible GitHub activities. Thus, the task construction used in the previous datasets is not applicable to \textsc{pres-github}. We decided to omit this dataset from the set of datasets used for creating personal event prediction tasks.

\paragraph{Full event sequence.}
In addition to the datasets containing prediction tasks described above, we also publish a version of each dataset that includes all personal and relational events for all users, without any assigned tasks, train/val/test splits, or pre-specified negative samples. This is intended to facilitate future works that may wish to generate other prediction tasks not covered in this paper.

\section{Experiments}

\label{sec:experiment}

\subsection{Relational event prediction tasks}

\begin{table}
  \caption{Performance results for relational event prediction tasks (all metrics are in percent).}
  \label{tab:results-rel}
  \setlength{\tabcolsep}{5pt}
  \resizebox{\textwidth}{!}{%
    \begin{tabular}{l|ccccc|ccccc}
      \toprule
      Method & \multicolumn{5}{c|}{\textsc{pres-brightkite}} & \multicolumn{5}{c}{\textsc{pres-gowalla}} \\
      Metric & MRR & Hits@5 & Hits@10 & Hits@50 & Hits@100 & MRR  & Hits@5 & Hits@10 & Hits@50 & Hits@100 \\
      \midrule
      \multicolumn{11}{l}{\textcolor{darkgray}{\textbf{Static graph models on relational event graph}}} \\
      GCN & 37.3{\scriptsize $\pm$0.8} & 50.8{\scriptsize $\pm$1.0} & 61.7{\scriptsize $\pm$0.9} & 83.2{\scriptsize $\pm$0.4} & 89.5{\scriptsize $\pm$0.3} & 40.3{\scriptsize $\pm$0.9} & 54.5{\scriptsize $\pm$0.9} & 65.8{\scriptsize $\pm$0.8} & 86.5{\scriptsize $\pm$0.4} & 92.0{\scriptsize $\pm$0.2} 
      \\
      GAT & 36.2{\scriptsize $\pm$1.4} & 48.7{\scriptsize $\pm$1.4} & 59.5{\scriptsize $\pm$1.2} & 81.4{\scriptsize $\pm$0.8} & 88.5{\scriptsize $\pm$0.6} & 40.7{\scriptsize $\pm$1.5} & 54.1{\scriptsize $\pm$1.6} & 64.9{\scriptsize $\pm$1.5} & 85.3{\scriptsize $\pm$1.3} & 91.1{\scriptsize $\pm$1.0} \\
      TConv & 40.2{\scriptsize $\pm$2.0} & 53.0{\scriptsize $\pm$2.4} & 63.5{\scriptsize $\pm$2.3} & 84.1{\scriptsize $\pm$1.1} & 90.0{\scriptsize $\pm$0.6} & \underline{47.4{\scriptsize $\pm$1.1}} & \underline{62.0{\scriptsize $\pm$1.1}} & \underline{72.1{\scriptsize $\pm$0.8}} & \underline{88.9{\scriptsize $\pm$0.3}} & \underline{93.1{\scriptsize $\pm$0.2}} \\
      \midrule
      \multicolumn{11}{l}{\textcolor{darkgray}{\textbf{Static graph models on relational event graph + sequence embedding from personal event data}}} \\
      GCN+S & {43.9{\scriptsize $\pm$0.7}} & {57.8{\scriptsize $\pm$0.8}} & {67.8{\scriptsize $\pm$0.8}} & \underline{86.5{\scriptsize $\pm$0.3}} & \underline{91.5{\scriptsize $\pm$0.1}} & {44.9{\scriptsize $\pm$1.0}} & {59.4{\scriptsize $\pm$1.1}} & {69.8{\scriptsize $\pm$1.0}} & {88.1{\scriptsize $\pm$0.5}} & {92.8{\scriptsize $\pm$0.3}} \\
      GAT+S & \underline{44.8{\scriptsize $\pm$1.1}} & \underline{58.5{\scriptsize $\pm$1.1}} & \underline{68.2{\scriptsize $\pm$1.1}} & {86.2{\scriptsize $\pm$0.5}} & {91.5{\scriptsize $\pm$0.4}} & {44.9{\scriptsize $\pm$0.9}} & {58.8{\scriptsize $\pm$0.6}} & {69.0{\scriptsize $\pm$0.4}} & {87.0{\scriptsize $\pm$0.4}} & {92.0{\scriptsize $\pm$0.5}} \\
      TConv+S & \textbf{47.4{\scriptsize $\pm$1.5}} & \textbf{61.6{\scriptsize $\pm$1.7}} & \textbf{71.4{\scriptsize $\pm$1.4}} & \textbf{88.2{\scriptsize $\pm$0.4}} & \textbf{92.5{\scriptsize $\pm$0.2}} & \textbf{49.9{\scriptsize $\pm$1.1}} & \textbf{64.6{\scriptsize $\pm$1.1}} & \textbf{74.4{\scriptsize $\pm$0.9}} & \textbf{90.0{\scriptsize $\pm$0.5}} & \textbf{93.8{\scriptsize $\pm$0.3}} \\
      \midrule
      \multicolumn{11}{l}{\textcolor{darkgray}{\textbf{Static graph models on relational event graph + personal event nodes}}} \\
      GCN\_RP & 8.7{\scriptsize $\pm$0.9} & 11.0{\scriptsize $\pm$1.2} & 15.7{\scriptsize $\pm$1.7} & 35.6{\scriptsize $\pm$3.7} & 49.8{\scriptsize $\pm$4.5} & 17.0{\scriptsize $\pm$0.9} & 22.1{\scriptsize $\pm$1.2} & 29.8{\scriptsize $\pm$1.6} & 56.4{\scriptsize $\pm$3.0} & 70.8{\scriptsize $\pm$2.9} \\
      GAT\_RP & 10.7{\scriptsize $\pm$1.0} & 13.5{\scriptsize $\pm$1.2} & 18.2{\scriptsize $\pm$1.4} & 35.6{\scriptsize $\pm$2.3} & 47.8{\scriptsize $\pm$2.8} & 14.9{\scriptsize $\pm$1.4} & 19.0{\scriptsize $\pm$1.6} & 25.8{\scriptsize $\pm$2.0} & 50.7{\scriptsize $\pm$3.1} & 66.2{\scriptsize $\pm$3.2} \\
      TConv\_RP & 15.8{\scriptsize $\pm$1.6} & 21.3{\scriptsize $\pm$2.3} & 29.2{\scriptsize $\pm$2.7} & 55.7{\scriptsize $\pm$3.2} & 70.4{\scriptsize $\pm$2.7} & 21.0{\scriptsize $\pm$1.5} & 28.2{\scriptsize $\pm$2.2} & 38.1{\scriptsize $\pm$2.7} & 67.4{\scriptsize $\pm$3.0} & 80.4{\scriptsize $\pm$2.4} \\
      \midrule
      \multicolumn{11}{l}{\textcolor{darkgray}{\textbf{Temporal graph models on relational event graph + personal event nodes}}} \\
      TGN & 12.2{\scriptsize $\pm$0.7} & 15.9{\scriptsize $\pm$0.9} & 23.5{\scriptsize $\pm$1.0} & 50.2{\scriptsize $\pm$1.3} & 63.5{\scriptsize $\pm$1.3} & 15.4{\scriptsize $\pm$2.6} & 20.6{\scriptsize $\pm$3.7} & 27.8{\scriptsize $\pm$4.6} & 51.8{\scriptsize $\pm$5.6} & 64.9{\scriptsize $\pm$5.4} \\
      DyRep & 7.1{\scriptsize $\pm$0.4} & 8.9{\scriptsize $\pm$0.6} & 13.7{\scriptsize $\pm$0.9} & 36.0{\scriptsize $\pm$1.7} & 50.7{\scriptsize $\pm$2.1} & 8.8{\scriptsize $\pm$1.0} & 11.2{\scriptsize $\pm$1.3} & 15.8{\scriptsize $\pm$1.7} & 34.8{\scriptsize $\pm$3.5} & 48.6{\scriptsize $\pm$5.1} \\
      TNCN  & 28.5{\scriptsize $\pm$1.3} & 34.5{\scriptsize $\pm$1.6} & 40.7{\scriptsize $\pm$1.9} & 62.6{\scriptsize $\pm$2.1} & 72.8{\scriptsize $\pm$1.8} & 25.1{\scriptsize $\pm$1.6} & 29.2{\scriptsize $\pm$2.0} & 33.7{\scriptsize $\pm$2.4} & 52.2{\scriptsize $\pm$2.5} & 63.4{\scriptsize $\pm$2.4} \\
      \bottomrule
    \end{tabular}%
  }
  \vspace{1em}
  \resizebox{\textwidth}{!}{%
    \begin{tabular}{l|ccccc|ccccc}
      \toprule
      Method & \multicolumn{5}{c|}{\textsc{pres-amazon-clothing}} & \multicolumn{5}{c}{\textsc{pres-amazon-electronics}} \\
      Metric & MRR & Hits@5 & Hits@10 & Hits@50 & Hits@100 & MRR  & Hits@5 & Hits@10 & Hits@50 & Hits@100 \\
      \midrule
      \multicolumn{11}{l}{\textcolor{darkgray}{\textbf{Static graph models on relational event graph}}} \\
      GCN & 6.1{\scriptsize $\pm$1.6} & 7.4{\scriptsize $\pm$2.1} & 10.0{\scriptsize $\pm$2.5} & 23.4{\scriptsize $\pm$3.0} & 35.3{\scriptsize $\pm$1.3} & 13.1{\scriptsize $\pm$0.6} & 15.9{\scriptsize $\pm$0.7} & 21.5{\scriptsize $\pm$0.6} & 45.9{\scriptsize $\pm$1.3} & 60.6{\scriptsize $\pm$1.6} \\
      GAT & 7.2{\scriptsize $\pm$2.5} & 7.8{\scriptsize $\pm$2.7} & 10.2{\scriptsize $\pm$2.9} & 23.8{\scriptsize $\pm$3.6} & 38.4{\scriptsize $\pm$1.9} & 13.2{\scriptsize $\pm$0.7} & 15.5{\scriptsize $\pm$0.9} & 20.7{\scriptsize $\pm$1.0} & 45.2{\scriptsize $\pm$1.2} & 61.0{\scriptsize $\pm$1.5} \\
      TConv & 4.2{\scriptsize $\pm$0.3} & 4.8{\scriptsize $\pm$0.7} & 7.7{\scriptsize $\pm$1.0} & 23.5{\scriptsize $\pm$1.6} & 35.5{\scriptsize $\pm$2.0} & 17.3{\scriptsize $\pm$0.3} & 24.1{\scriptsize $\pm$0.4} & 34.3{\scriptsize $\pm$0.5} & \underline{63.0{\scriptsize $\pm$0.6}} & \underline{73.7{\scriptsize $\pm$0.4}} \\
      \midrule
      \multicolumn{11}{l}{\textcolor{darkgray}{\textbf{Static graph models on relational event graph + sequence embedding from personal event data}}} \\
      GCN+S & 4.5{\scriptsize $\pm$0.3} & 5.5{\scriptsize $\pm$0.6} & 9.3{\scriptsize $\pm$0.8} & {29.0{\scriptsize $\pm$0.5}} & {40.4{\scriptsize $\pm$0.4}} & {14.7{\scriptsize $\pm$0.5}} & {19.1{\scriptsize $\pm$0.8}} & {27.2{\scriptsize $\pm$1.5}} & {57.9{\scriptsize $\pm$1.9}} & {70.6{\scriptsize $\pm$1.5}} \\
      GAT+S & {7.7{\scriptsize $\pm$2.1}} & {8.5{\scriptsize $\pm$2.1}} & {12.0{\scriptsize $\pm$1.8}} & {31.3{\scriptsize $\pm$1.3}} & \underline{46.3{\scriptsize $\pm$0.6}} & 14.4{\scriptsize $\pm$1.7} & 16.7{\scriptsize $\pm$1.8} & 21.6{\scriptsize $\pm$1.4} & 43.6{\scriptsize $\pm$2.5} & 58.4{\scriptsize $\pm$3.4} \\
      TConv+S & 7.5{\scriptsize $\pm$0.5} & \underline{9.6{\scriptsize $\pm$0.7}} & \underline{14.7{\scriptsize $\pm$0.9}} & \textbf{37.9{\scriptsize $\pm$0.9}} & \textbf{50.6{\scriptsize $\pm$0.4}} & \underline{22.3{\scriptsize $\pm$1.6}} & \underline{32.1{\scriptsize $\pm$2.4}} & \textbf{44.5{\scriptsize $\pm$2.3}} & \textbf{70.1{\scriptsize $\pm$1.1}} & \textbf{78.1{\scriptsize $\pm$0.8}} \\
      \midrule
      \multicolumn{11}{l}{\textcolor{darkgray}{\textbf{Static graph models on relational event graph + personal event nodes}}} \\
      GCN\_RP & \textbf{8.7{\scriptsize $\pm$1.4}} & {9.2{\scriptsize $\pm$1.4}} & 10.5{\scriptsize $\pm$1.4} & 18.1{\scriptsize $\pm$1.2} & 25.8{\scriptsize $\pm$2.5} & 7.5{\scriptsize $\pm$0.6} & 8.3{\scriptsize $\pm$0.6} & 10.9{\scriptsize $\pm$0.8} & 21.8{\scriptsize $\pm$2.3} & 29.0{\scriptsize $\pm$3.2} \\
      GAT\_RP & 6.5{\scriptsize $\pm$1.0} & 7.9{\scriptsize $\pm$1.0} & {10.9{\scriptsize $\pm$1.4}} & 25.7{\scriptsize $\pm$3.2} & 39.8{\scriptsize $\pm$3.7} & {15.5{\scriptsize $\pm$0.5}} & 17.2{\scriptsize $\pm$0.4} & 20.5{\scriptsize $\pm$0.7} & 35.5{\scriptsize $\pm$3.0} & 46.9{\scriptsize $\pm$3.6} \\
      TConv\_RP & 4.7{\scriptsize $\pm$0.3} & 5.7{\scriptsize $\pm$0.4} & 8.4{\scriptsize $\pm$0.7} & 22.9{\scriptsize $\pm$0.9} & 34.3{\scriptsize $\pm$0.9} & 13.7{\scriptsize $\pm$0.7} & 18.2{\scriptsize $\pm$0.9} & 25.5{\scriptsize $\pm$1.2} & 50.8{\scriptsize $\pm$1.5} & 64.3{\scriptsize $\pm$1.1} \\
      \midrule
      \multicolumn{11}{l}{\textcolor{darkgray}{\textbf{Temporal graph models on relational event graph + personal event nodes}}} \\
      TGN & 3.5{\scriptsize $\pm$0.5} & 4.1{\scriptsize $\pm$1.2} & 6.9{\scriptsize $\pm$1.2} & 23.7{\scriptsize $\pm$0.8} & 39.0{\scriptsize $\pm$1.3} & 13.8{\scriptsize $\pm$0.3} & {19.2{\scriptsize $\pm$0.6}} & {26.4{\scriptsize $\pm$0.9}} & {48.8{\scriptsize $\pm$0.9}} & {61.5{\scriptsize $\pm$0.7}} \\
      DyRep & 2.9{\scriptsize $\pm$0.6} & 3.0{\scriptsize $\pm$1.1} & 5.8{\scriptsize $\pm$1.6} & 22.8{\scriptsize $\pm$2.6} & 39.6{\scriptsize $\pm$2.4} & 6.8{\scriptsize $\pm$0.7} & 8.9{\scriptsize $\pm$1.0} & 13.8{\scriptsize $\pm$1.3} & 33.4{\scriptsize $\pm$2.3} & 47.5{\scriptsize $\pm$3.0} \\
      TNCN & \underline{8.2{\scriptsize $\pm$1.8}} & \textbf{11.5{\scriptsize $\pm$2.7}} & \textbf{16.3{\scriptsize $\pm$2.6}} & \underline{31.6{\scriptsize $\pm$3.2}} & 39.3{\scriptsize $\pm$3.2} & \textbf{25.5{\scriptsize $\pm$1.8}} & \textbf{32.2{\scriptsize $\pm$1.8}} & \underline{38.4{\scriptsize $\pm$1.5}} & 56.2{\scriptsize $\pm$1.7} & 63.3{\scriptsize $\pm$2.3} \\
      \bottomrule
    \end{tabular}%
  }
\end{table}

\textbf{Experiment setup.}
We perform relational event prediction experiments on all five PRES datasets, following the task setup described earlier. We evaluate several sets of baseline methods:
\begin{enumerate}
\setlength\itemsep{0.2pt}
\setlength\topsep{0.2pt}
    \item In the first set, we use only relational event data. We construct a user graph where edges represent relational events between two users, ignoring timestamp information. We then run static graph methods, GCN \citep{kipf2017semi}, GAT \citep{velivckovic2018graph}, and Graph Transformer (TConv) \cite{shi2021masked} on this graph.


    \item In the second set, we use a biderectional transformer sequence model (BERT) \citep{devlin2019bert}, to encode each user's last 100 personal events from the training set. The resulting user embedding is added as input to the GCN, GAT, and TConv models from the first set, denoted as GCN+S, GAT+S, and TConv+S, respectively.

    \item In the third set, we convert each unique personal event into a node and add it to the user graph from the first set, creating edges between users and their personal event nodes. As in the second set, we use only the last 100 personal events per user. We then run GCN, GAT, and TConv on this graph, denoted as GCN\_RP, GAT\_RP and TConv\_RP.
    
    \item Lastly, based on the graph containing user and personal event nodes from the third set, we add timestamp information to construct a temporal graph. For datasets that lack timestamps for relational events, we inject these events randomly into the sequence of personal events. We then run temporal graph models, TGN \citep{tgn_icml_grl2020}, DyRep \citep{trivedi2019dyrep}, and TNCN \citep{zhang2406efficient} on this  graph.
\end{enumerate}


\looseness=-1
The sequence model for capturing personal events in the second set is designed as a masked token prediction task using a BERT model with a masking probability of 0.3. A key benefit of using transformer-based models is flexibility in event tokenization. In \textsc{pres-brightkite} and \textsc{pres-gowalla}, personal events are 8-character geohash strings (e.g., {9q8yyk8y|9q8vzj5b|9q8vyzwk}). Since geohashes encode hierarchical geographic information, we apply hierarchical tokenization by splitting each into four two-character tokens with added prefixes (e.g., {gh12-9q}, {gh34-8y}, {gh12-yk}, {gh12-8y}). This roughly mimics hierarchical location modeling, such as identifying continent, country, city, and neighborhood. For the \textsc{pres-amazon} datasets, we apply similar tokenization by splitting each event into three product tokens and one rating token. We do not apply token splitting for \textsc{pres-github}.

For performance evaluation, following prior benchmarks \citep{huang2023temporal, gastinger2024tgb, yi2025tgb}, we use ranking-based metrics: Mean Reciprocal Rank (MRR) and Hits@k, evaluated at various $k$ depending on the number of negative samples. Each baseline is run five times with different random seeds, and we report the mean and standard deviation of the results.

\looseness=-1
\textbf{Experiment results.}
Table~\ref{tab:results-rel} and Table~\ref{tab:github} show the experiment results, with additional results available in Appendix \ref{sec:ap-experiments}. In each table, bold numbers indicate the best-performing model on a given metric, and underlined numbers indicate the second best. As each dataset has its own characteristics, the results vary across datasets. However, there are some emerging patterns in the results that we highlight below.

\begin{wraptable}{r}{0.6\textwidth}
\vspace{-4mm}
  \centering
  \caption{Relational event predictions on \textsc{pres-github}.}
  \label{tab:github}
  \setlength{\tabcolsep}{6pt}
  \resizebox{\linewidth}{!}{%
    \begin{tabular}{l|ccccc}
      \toprule
      Method & MRR & Hits@3  & Hits@5 & Hits@10 & Hits@30  \\
      \midrule
      GCN   & 54.0{\scriptsize $\pm$7.2} & 62.9{\scriptsize $\pm$7.5} & 69.6{\scriptsize $\pm$5.1} & 75.2{\scriptsize $\pm$2.3} & 80.1{\scriptsize $\pm$0.4} \\
      GAT   & 69.3{\scriptsize $\pm$3.1} & 73.6{\scriptsize $\pm$2.2} & 76.1{\scriptsize $\pm$1.3} & 78.1{\scriptsize $\pm$0.4} & 80.4{\scriptsize $\pm$0.3} \\
      TConv & 54.5{\scriptsize $\pm$1.9} & 62.8{\scriptsize $\pm$2.1} & 69.2{\scriptsize $\pm$1.5} & 74.6{\scriptsize $\pm$0.7} & 78.0{\scriptsize $\pm$0.2} \\
      \midrule
      GCN+S   & \underline{70.8{\scriptsize $\pm$0.8}} & \underline{75.1{\scriptsize $\pm$0.2}} & \underline{76.9{\scriptsize $\pm$0.0}} & \underline{78.5{\scriptsize $\pm$0.1}} & \underline{80.9{\scriptsize $\pm$0.4}} \\
      GAT+S   & \textbf{74.2{\scriptsize $\pm$0.6}} & \textbf{77.0{\scriptsize $\pm$0.4}} & \textbf{78.7{\scriptsize $\pm$0.3}} & \textbf{80.6{\scriptsize $\pm$0.1}} & \textbf{84.5{\scriptsize $\pm$0.2}} \\
      TConv+S & 62.7{\scriptsize $\pm$1.0} & 70.6{\scriptsize $\pm$0.7} & 74.7{\scriptsize $\pm$0.4} & 77.4{\scriptsize $\pm$0.2} & 78.8{\scriptsize $\pm$0.1} \\
      \midrule
      GCN\_RP  & 22.3{\scriptsize $\pm$2.6} & 23.3{\scriptsize $\pm$2.9} & 28.8{\scriptsize $\pm$3.1} & 37.8{\scriptsize $\pm$3.3} & 57.5{\scriptsize $\pm$3.5} \\
      GAT\_RP  & 33.1{\scriptsize $\pm$4.1} & 35.7{\scriptsize $\pm$5.4} & 43.9{\scriptsize $\pm$5.6} & 57.2{\scriptsize $\pm$4.5} & 76.7{\scriptsize $\pm$0.7} \\
      TConv\_RP & 33.6{\scriptsize $\pm$1.5} & 39.2{\scriptsize $\pm$2.2} & 50.1{\scriptsize $\pm$2.2} & 64.2{\scriptsize $\pm$1.4} & 76.3{\scriptsize $\pm$0.2} \\
      \midrule
      TGN     & \multicolumn{5}{c}{Out of GPU Memory} \\
      DyRep   & \multicolumn{5}{c}{Out of GPU Memory} \\
      TNCN   & \multicolumn{5}{c}{Out of GPU Memory} \\      
      \bottomrule
    \end{tabular}%
  }
\end{wraptable}

\begin{itemize}
\setlength\itemsep{0.2pt}
\setlength\topsep{0.2pt}




    \item The graph models with personal event sequence embeddings---GCN+S, GAT+S, and especially TConv+S---consistently perform well across all datasets and metrics. On \textsc{pres-brightkite} and \textsc{pres-gowalla}, TConv+S outperforms all models across all metrics, often by a wide margin, especially compared to methods that encode personal events as nodes. On \textsc{pres-github}, GAT+S shines, outperforming all methods on all metrics.

    \looseness=-1
    \item For the \textsc{amazon} datasets, TConv+S remains dominant on Hits@k metrics with larger $k$, achieving the best results at Hits@10, Hits@50, and Hits@100 on \textsc{pres-amazon-clothing}, as well as Hits@50 and Hits@100 on \textsc{pres-amazon-electronics}, while holding second place at other $k$ values. At these lower $k$, where TConv+S ranks second, the best model is TNCN, a temporal graph method. TNCN, however, does not perform well on Hits@k metrics with larger $k$.


    \item With the exception of TNCN on the \textsc{amazon} datasets with small $k$ metrics, the performance of temporal graph methods (TGN, DyRep, and TNCN) using graphs with personal event nodes is noticeably lower compared to static graph models with sequence embeddings. For the large \textsc{pres-github} dataset, they encounter GPU out-of-memory errors, even when using small batch sizes.


    \item In nearly all cases, converting personal events into nodes and adding them to the relational event graph is less effective than modeling personal events with a sequence model and using the user's sequence embedding as an additional node features to the relational event graph. The performance of the “\_RP” versions of static graph models is lower than that of the corresponding “+S” versions across nearly all datasets and metrics, with a few exceptions on the \textsc{amazon} datasets for Hits@k metrics with lower $k$.
   


    
\end{itemize}

\looseness=-1
Even though GCN+S, GAT+S, and TConv+S perform relational event prediction in two stages, first generating user embeddings from personal event sequences and then incorporating them into the graph learning process, they still perform well across datasets. In contrast, TGN, DyRep, and TNCN use a single-step approach that directly integrates temporal dynamics but operates on graphs where personal events are represented as nodes. These differences highlight an {opportunity for future exploration} of how best to represent the temporal dynamics of personal events within a user while jointly modeling the full structure that includes user-to-user relational events, in an {end-to-end fashion}.

\subsection{Personal event prediction tasks}

\begin{table}
  \caption{Performance results for personal event prediction tasks (all metrics are in percent).}
  \label{tab:results-pers}
  \setlength{\tabcolsep}{5pt}
  \resizebox{\textwidth}{!}{%
    \begin{tabular}{l|ccccc|ccccc}
      \toprule
      Method & \multicolumn{5}{c|}{\textsc{pres-brightkite}} & \multicolumn{5}{c}{\textsc{pres-gowalla}} \\
      Metric & MRR & Hits@3  & Hits@5  & Hits@10  & Hits@50 & MRR & Hits@3 & Hits@5 & Hits@10 & Hits@50 \\
      \midrule
      \multicolumn{11}{l}{\textcolor{darkgray}{\textbf{Sequential models}}} \\
      BERT        & \underline{34.2{\scriptsize $\pm$0.1}} & \underline{35.6{\scriptsize $\pm$0.2}} & \underline{37.4{\scriptsize $\pm$0.2}} & \underline{40.1{\scriptsize $\pm$0.2}} & 50.1{\scriptsize $\pm$0.3} & 15.3{\scriptsize $\pm$0.2} & 15.7{\scriptsize $\pm$0.3} & 18.6{\scriptsize $\pm$0.3} & 23.4{\scriptsize $\pm$0.3} & 43.1{\scriptsize $\pm$0.3} \\
      BERT-n2v-p & 33.8{\scriptsize $\pm$0.1} & 35.1{\scriptsize $\pm$0.2} & 36.9{\scriptsize $\pm$0.2} & 39.6{\scriptsize $\pm$0.2} & 49.8{\scriptsize $\pm$0.2} & 14.4{\scriptsize $\pm$0.2} & 14.8{\scriptsize $\pm$0.2} & 17.7{\scriptsize $\pm$0.2} & 22.6{\scriptsize $\pm$0.2} & 42.4{\scriptsize $\pm$0.2} \\
      BERT-n2v-i & \textbf{34.4{\scriptsize $\pm$0.1}} & \textbf{35.9{\scriptsize $\pm$0.1}} & \textbf{37.6{\scriptsize $\pm$0.1}} & \textbf{40.3{\scriptsize $\pm$0.1}} & 50.3{\scriptsize $\pm$0.2} & 15.0{\scriptsize $\pm$0.3} & 15.4{\scriptsize $\pm$0.3} & 18.3{\scriptsize $\pm$0.3} & 23.2{\scriptsize $\pm$0.3} & 42.7{\scriptsize $\pm$0.3} \\
      \midrule
      \multicolumn{11}{l}{\textcolor{darkgray}{\textbf{Graph models on personal event only graph}}} \\
      GCN       & 24.9{\scriptsize $\pm$1.2} & 27.1{\scriptsize $\pm$1.5} & 31.8{\scriptsize $\pm$1.7} & 38.8{\scriptsize $\pm$1.9} & \underline{55.8{\scriptsize $\pm$1.0}} & \underline{28.2{\scriptsize $\pm$3.2}} & \underline{29.7{\scriptsize $\pm$3.3}} & \underline{34.2{\scriptsize $\pm$3.0}} & \underline{41.5{\scriptsize $\pm$2.3}} & \underline{63.8{\scriptsize $\pm$0.9}} \\
      GAT       & 19.0{\scriptsize $\pm$1.4} & 20.3{\scriptsize $\pm$1.7} & 24.9{\scriptsize $\pm$1.9} & 32.1{\scriptsize $\pm$2.0} & 52.3{\scriptsize $\pm$1.6} & 15.4{\scriptsize $\pm$1.2} & 15.4{\scriptsize $\pm$1.4} & 20.0{\scriptsize $\pm$1.5} & 28.4{\scriptsize $\pm$1.6} & 59.3{\scriptsize $\pm$1.1} \\
      TGN       & 23.5{\scriptsize $\pm$0.2} & 24.5{\scriptsize $\pm$0.3} & 28.9{\scriptsize $\pm$0.4} & 37.1{\scriptsize $\pm$0.8} & 54.5{\scriptsize $\pm$1.1} & 10.7{\scriptsize $\pm$0.4} & 10.6{\scriptsize $\pm$0.6} & 14.4{\scriptsize $\pm$1.1} & 21.5{\scriptsize $\pm$1.8} & 42.7{\scriptsize $\pm$4.9} \\
      DyRep     & 19.8{\scriptsize $\pm$2.9} & 21.4{\scriptsize $\pm$3.2} & 26.5{\scriptsize $\pm$2.5} & 35.4{\scriptsize $\pm$1.6} & \textbf{57.2{\scriptsize $\pm$1.7}} & 7.4{\scriptsize $\pm$0.6} & 6.4{\scriptsize $\pm$0.8} & 10.0{\scriptsize $\pm$1.0} & 17.8{\scriptsize $\pm$1.0} & 42.9{\scriptsize $\pm$2.1} \\
      \midrule
      \multicolumn{11}{l}{\textcolor{darkgray}{\textbf{Graph models on personal and relational event graph}}} \\
      GCN\_PR      & 25.4{\scriptsize $\pm$1.2} & 27.5{\scriptsize $\pm$1.3} & 31.9{\scriptsize $\pm$1.5} & 38.2{\scriptsize $\pm$1.7} & 54.7{\scriptsize $\pm$1.6} & \textbf{30.3{\scriptsize $\pm$5.1}} & \textbf{32.0{\scriptsize $\pm$5.4}} & \textbf{36.8{\scriptsize $\pm$5.2}} & \textbf{44.2{\scriptsize $\pm$4.4}} & \textbf{65.4{\scriptsize $\pm$1.2}} \\
      GAT\_PR      & 18.8{\scriptsize $\pm$0.5} & 20.3{\scriptsize $\pm$0.6} & 25.2{\scriptsize $\pm$0.6} & 32.9{\scriptsize $\pm$0.6} & 53.3{\scriptsize $\pm$0.6} & 16.0{\scriptsize $\pm$0.6} & 16.0{\scriptsize $\pm$0.7} & 20.5{\scriptsize $\pm$0.8} & 28.6{\scriptsize $\pm$0.9} & 59.2{\scriptsize $\pm$0.9} \\
      TGN\_PR      & 29.5{\scriptsize $\pm$2.3} & 33.5{\scriptsize $\pm$2.2} & 35.3{\scriptsize $\pm$2.2} & 36.0{\scriptsize $\pm$2.4} & 36.1{\scriptsize $\pm$2.4} & 14.0{\scriptsize $\pm$2.0} & 15.7{\scriptsize $\pm$2.2} & 17.0{\scriptsize $\pm$2.4} & 17.9{\scriptsize $\pm$2.5} & 18.2{\scriptsize $\pm$2.6} \\
      DyRep\_PR    & 23.4{\scriptsize $\pm$2.7} & 27.5{\scriptsize $\pm$3.5} & 30.5{\scriptsize $\pm$3.2} & 32.7{\scriptsize $\pm$4.2} & 33.3{\scriptsize $\pm$4.8} & 10.5{\scriptsize $\pm$1.4} & 11.4{\scriptsize $\pm$1.8} & 12.4{\scriptsize $\pm$2.2} & 13.1{\scriptsize $\pm$2.8} & 13.6{\scriptsize $\pm$3.4} \\
      \bottomrule
    \end{tabular}%
  }
  \vspace{1em}
  \resizebox{\textwidth}{!}{%
    \begin{tabular}{l|ccccc|ccccc}
      \toprule
      Method & \multicolumn{5}{c|}{\textsc{pres-amazon-clothing}} & \multicolumn{5}{c}{\textsc{pres-amazon-electronics}} \\
      Metric & MRR & Hits@3  & Hits@5  & Hits@10  & Hits@50 & MRR & Hits@3 & Hits@5 & Hits@10 & Hits@50 \\
      \midrule
      \multicolumn{11}{l}{\textcolor{darkgray}{\textbf{Sequential models}}} \\
      BERT     &  3.3{\scriptsize $\pm$0.0} &  2.3{\scriptsize $\pm$0.0} &  3.5{\scriptsize $\pm$0.1} &  6.1{\scriptsize $\pm$0.1} & 22.4{\scriptsize $\pm$0.2} &  8.1{\scriptsize $\pm$0.2} &  7.7{\scriptsize $\pm$0.2} & 10.7{\scriptsize $\pm$0.3} & 16.1{\scriptsize $\pm$0.3} & 38.1{\scriptsize $\pm$0.2} \\
      BERT+n2v-p & 3.4{\scriptsize $\pm$0.0} &  2.4{\scriptsize $\pm$0.0} &  3.6{\scriptsize $\pm$0.0} &  6.2{\scriptsize $\pm$0.1} & 22.7{\scriptsize $\pm$0.2} &  8.1{\scriptsize $\pm$0.1} &  7.7{\scriptsize $\pm$0.1} & 10.7{\scriptsize $\pm$0.2} & 16.1{\scriptsize $\pm$0.2} & 38.2{\scriptsize $\pm$0.1} \\
      BERT+n2v-i & 3.3{\scriptsize $\pm$0.0} &  2.3{\scriptsize $\pm$0.0} &  3.5{\scriptsize $\pm$0.1} &  6.1{\scriptsize $\pm$0.1} & 22.6{\scriptsize $\pm$0.2} &  8.1{\scriptsize $\pm$0.1} &  7.8{\scriptsize $\pm$0.1} & 10.7{\scriptsize $\pm$0.0} & 16.1{\scriptsize $\pm$0.1} & 38.0{\scriptsize $\pm$0.2} \\
      \midrule
      \multicolumn{11}{l}{\textcolor{darkgray}{\textbf{Graph models on personal event only graph}}} \\
      GCN    & \underline{10.8{\scriptsize $\pm$1.7}} & \underline{11.2{\scriptsize $\pm$2.0}} & 14.1{\scriptsize $\pm$1.7} & 19.0{\scriptsize $\pm$1.1} & 32.8{\scriptsize $\pm$0.5} & 13.3{\scriptsize $\pm$2.0} & 13.3{\scriptsize $\pm$2.5} & 18.4{\scriptsize $\pm$2.9} & 27.6{\scriptsize $\pm$3.1} & \underline{55.6{\scriptsize $\pm$0.9}} \\
      GAT    &  3.5{\scriptsize $\pm$0.0} &  2.6{\scriptsize $\pm$0.0} &  4.0{\scriptsize $\pm$0.1} &  7.1{\scriptsize $\pm$0.1} & 22.7{\scriptsize $\pm$0.2} &  7.4{\scriptsize $\pm$0.4} &  6.4{\scriptsize $\pm$0.4} &  9.6{\scriptsize $\pm$0.5} & 16.1{\scriptsize $\pm$0.7} & 44.0{\scriptsize $\pm$0.8} \\
      TGN    &  9.3{\scriptsize $\pm$0.9} &  8.0{\scriptsize $\pm$0.8} & 12.8{\scriptsize $\pm$2.2} & 25.4{\scriptsize $\pm$6.8} & \textbf{44.4{\scriptsize $\pm$3.4}} & \underline{16.2{\scriptsize $\pm$1.6}} & \textbf{17.4{\scriptsize $\pm$2.0}} & \underline{22.6{\scriptsize $\pm$1.8}} & \underline{30.8{\scriptsize $\pm$1.2}} & 54.2{\scriptsize $\pm$0.8} \\
      DyRep  &  8.9{\scriptsize $\pm$1.3} &  8.1{\scriptsize $\pm$2.4} & 13.5{\scriptsize $\pm$4.1} & 25.1{\scriptsize $\pm$6.7} & \underline{43.5{\scriptsize $\pm$5.7}} & 11.4{\scriptsize $\pm$0.4} & 10.6{\scriptsize $\pm$0.6} & 15.3{\scriptsize $\pm$0.8} & 25.8{\scriptsize $\pm$1.0} & 55.1{\scriptsize $\pm$0.8} \\
      \midrule
      \multicolumn{11}{l}{\textcolor{darkgray}{\textbf{Graph models on personal and relational event graph}}} \\
      GCN\_PR   & \textbf{10.9{\scriptsize $\pm$1.3}} & \textbf{11.4{\scriptsize $\pm$1.5}} & \underline{14.3{\scriptsize $\pm$1.3}} & 19.1{\scriptsize $\pm$0.8} & 32.8{\scriptsize $\pm$0.6} & \textbf{16.6{\scriptsize $\pm$1.5}} & \underline{17.3{\scriptsize $\pm$1.9}} & 22.2{\scriptsize $\pm$2.1} & 30.6{\scriptsize $\pm$2.0} & \textbf{56.3{\scriptsize $\pm$0.8}} \\
      GAT\_PR   &  3.5{\scriptsize $\pm$0.1} &  2.6{\scriptsize $\pm$0.1} &  4.0{\scriptsize $\pm$0.1} &  7.1{\scriptsize $\pm$0.2} & 22.5{\scriptsize $\pm$0.2} &  8.0{\scriptsize $\pm$0.3} &  7.2{\scriptsize $\pm$0.4} & 10.5{\scriptsize $\pm$0.5} & 17.2{\scriptsize $\pm$0.7} & 44.8{\scriptsize $\pm$0.6} \\
      TGN\_PR   &  9.3{\scriptsize $\pm$0.7} &  7.8{\scriptsize $\pm$1.1} & 13.9{\scriptsize $\pm$3.3} & \underline{30.4{\scriptsize $\pm$3.7}} & 41.7{\scriptsize $\pm$1.9} & 15.6{\scriptsize $\pm$0.6} & 16.8{\scriptsize $\pm$0.8} & \textbf{22.8{\scriptsize $\pm$0.8}} & \textbf{32.7{\scriptsize $\pm$0.6}} & 54.0{\scriptsize $\pm$1.0} \\
      DyRep\_PR & 10.5{\scriptsize $\pm$0.2} &  9.7{\scriptsize $\pm$0.5} & \textbf{17.1{\scriptsize $\pm$0.6}} & \textbf{32.9{\scriptsize $\pm$0.3}} & 41.3{\scriptsize $\pm$0.4} & 14.3{\scriptsize $\pm$0.5} & 15.0{\scriptsize $\pm$0.7} & 21.2{\scriptsize $\pm$0.9} & 32.3{\scriptsize $\pm$1.4} & 53.3{\scriptsize $\pm$2.8} \\
      \bottomrule
    \end{tabular}%
  }
\end{table}

\textbf{Experiment setup.}
We perform personal event prediction experiments on all PRES datasets except \textsc{pres-github}. In these experiments, we evaluate several sets of baseline methods:

\begin{enumerate}
\setlength\itemsep{0.3pt}
\setlength\topsep{0.3pt}

    \item The first model is a sequential model that uses only personal event data. We use a bidirectional transformer  architecture (BERT) with a prediction head to compute the likelihood of a user having a particular personal event in the future. For each user, we use the last 100 personal events in the training set to predict the likelihood of future events.

    \item In the second set, we use node2vec \citep{grover2016node2vec} to learn the graph structure of relational events and generate a graph embedding for each user. We then incorporate the embedding into the BERT sequence model. We evaluate two versions of the model: (a) incorporating the graph embedding post transformer module and before the prediction head (\textsc{BERT-n2v-p}), and (b) using the embedding as a special input token to the transformer module (\textsc{BERT-n2v-i}).

    \item In the third set, we use graph-based models on personal event–only data by creating a bipartite graph of user nodes and personal event nodes, based on the last 100 personal events per user. We run both static graph models (GCN and GAT) and temporal graph models (TGN and DyRep) on this graph.

    \item In the last set, we augment the graph in the third set with relational event data by adding relational event edges between users. We then run GCN, GAT, TGN, and DyRep on this graph, denoted as GCN\_PR, GAT\_PR, TGN\_PR, and DyRep\_PR, respectively.
\end{enumerate}


Similar to the sequence embedding used in relational event prediction tasks, we apply split tokenization for the BERT model in personal event prediction to allow more flexibility in  modeling events. We use the same tokenization scheme for each dataset as described earlier. For evaluation, we report MRR and Hits@k at various values of $k$. Each baseline is run five times with different random seeds, and we report the mean and standard deviation of the results.

\textbf{Experiment results.}
Table~\ref{tab:results-pers} shows the results for the personal event prediction task. As in the relational event task, results vary across datasets due to their unique characteristics, with even more variations in this setting. We discuss some of the results as follows.

\begin{itemize}
\setlength\itemsep{0.3pt}
\setlength\topsep{0.3pt}


    \item The sequence models perform well on \textsc{pres-brightkite} across all metrics. The base BERT model, which uses only personal event data, already shows strong performance. Adding relational event node2vec embeddings may either improve or degrade performance. In \textsc{pres-brightkite}, adding the embedding after the transformer module reduces performance, while using it as a special input token improves it. However, the changes are relatively minor but sufficient to make \textsc{BERT-n2v-i} the best-performing model on \textsc{pres-brightkite}. Similar minimal changes are observed in other datasets.

    \looseness=-1
    \item The static graph model, GCN in particular, performs surprisingly well on \textsc{pres-gowalla}. The best performance is achieved by the GCN\_PR model, which is trained on data containing both personal and relational events in a graph with user nodes and personal event nodes. GCN\_PR also performs relatively well on \textsc{pres-amazon-clothing} and \textsc{pres-amazon-electronics}. However, the GAT-based models perform noticeably worse than their GCN counterparts.

    \item The temporal graph models perform relatively well on the \textsc{pres-amazon-clothing} and \textsc{pres-amazon-electronics} datasets, particularly on the Hits@5 and Hits@10 metrics. TGN and DyRep perform better on graphs that include both personal and relational events. A notable exception is the Hits@50 metric.
    
\end{itemize}

The results show that there is no single model that consistently performs best across all datasets. Some models work well on certain datasets but not on others. The only consistent pattern is that the best-performing models usually use both personal and relational events. This opens up opportunities for designing better models that can effectively integrate both types of information.

\subsection{Hypothesis on native end-to-end model architectures for PRES datasets}

In the experiment results described above, we have demonstrated that current models leave notable room for performance improvements in both personal and relational prediction tasks. We hypothesize that a significant performance improvement can be achieved by end-to-end models trained simultaneously on both personal and relational event data. There are many ways to design such models; below are some examples:
\begin{enumerate}
    \item A possible strategy is to start with a user-level Transformer architecture that takes the user event sequence as input, combined with a message-passing architecture that enables information transfer between two users connected by a relational event and propagates that information across the neighbors via a Graph Neural Network (GNN)-like architecture. This design capitalizes on the strengths of both the Transformer and GNN architectures. Transformer architectures have been shown to be the best model for handling sequential information, whereas GNN architectures excel in transferring information across connected users.
    \item Another possible architecture candidate is to also start from the Transformer architecture for the sequence, where each sample represents a user event sequence, and allow the attention mechanism to propagate toward other user sequences that are connected via a relational event, in addition to the attention mechanism inside each user/sequence.
\end{enumerate}
There could also be multiple other promising architectural paradigms for developing a fully end-to-end model. We leave the exploration of these alternative approaches for future work.

\section{Conclusions and Limitations}

In this work, we aim to advance user event modeling by introducing a unified framework that captures both personal and relational events. We curate and release a collection of public datasets with corresponding prediction tasks, all aligned under a formalization that integrates both event types to provide a more complete view of user behavior. Through empirical evaluation, we demonstrate that models leveraging both event types consistently outperform those using only one. We also show that existing methods, originally developed for either sequential or relational data, even with some adaptations to handle both (e.g., temporal graph models), are less effective across many of our prediction tasks. These findings highlight the need for further study of unified user event modeling.

A key challenge in this work is in the dataset curation process, as many public datasets have already been collapsed into either graph-only or sequence-only formats, often discarding personal or relational events in the process. While we were able to gather and unify a set of datasets that include both event types, they may not fully capture the diversity and complexity of user event modeling across domains. Another limitation is that our current formulation does not support event-level or user-level features, presenting an opportunity for future work to extend the framework toward feature-aware modeling.

\bibliographystyle{unsrtnat}
\bibliography{ref}

@article{purificato2024user,
  title={User Modeling and User Profiling: A Comprehensive Survey},
  author={Purificato, Erasmo and Boratto, Ludovico and De Luca, Ernesto William},
  journal={arXiv preprint arXiv:2402.09660},
  year={2024}
}

@article{martin2021survey,
  title={A survey for user behavior analysis based on machine learning techniques: current models and applications},
  author={Mart{\'\i}n, Alejandro Garc{\'\i}a and Gonz{\'a}lez, Raquel Mart{\'\i}nez and Garc{\'\i}a, Andr{\'e}s and Villarrubia, Gabriel},
  journal={Applied Intelligence},
  volume={51},
  pages={8110--8127},
  year={2021},
  publisher={Springer}
}

@inproceedings{deng2024advances,
  title={Advances in Human Event Modeling: From Graph Neural Networks to Language Models},
  author={Deng, Songgaojun and de Rijke, Maarten and Ning, Yue},
  booktitle={Proceedings of the 30th ACM SIGKDD Conference on Knowledge Discovery and Data Mining},
  pages={6459--6469},
  year={2024}
}

@article{he2023survey,
  title={A Survey on User Behavior Modeling in Recommender Systems},
  author={He, Zhicheng and Liu, Weiwen and Guo, Wei and Qin, Jiarui and Zhang, Yingxue and Hu, Yaochen and Tang, Ruiming},
  journal={arXiv preprint arXiv:2302.11087},
  year={2023}
}

@article{zeng2019user,
  title={User behaviour modeling, recommendations, and purchase prediction during shopping festivals},
  author={Zeng, Meng and Cao, Hong and Chen, Min and Li, Yujie},
  journal={Electronic Markets},
  volume={29},
  number={2},
  pages={205--217},
  year={2019},
  publisher={Springer}
}

@article{abdelhafez2013survey,
  title={A survey of user modelling in social media websites},
  author={Abdel-Hafez, Ahmed and Xu, Yanchun},
  journal={Computer and Information Science},
  volume={6},
  number={4},
  pages={59--71},
  year={2013},
  publisher={Canadian Center of Science and Education}
}

@article{piao2018inferring,
  title={Inferring user interests in microblogging social networks: A survey},
  author={Piao, Guangyuan and Breslin, John G},
  journal={User Modeling and User-Adapted Interaction},
  volume={28},
  number={3},
  pages={277--329},
  year={2018},
  publisher={Springer}
}

@inproceedings{fang2023snapchat,
  title={General-Purpose User Modeling with Behavioral Logs: A Snapchat Case Study},
  author={Fang, Qixiang and Zhou, Zhihan and Barbieri, Francesco and Liu, Yozen and Neves, Leonardo and Nguyen, Dong and Oberski, Daniel L and Bos, Maarten W and Dotsch, Ron},
  booktitle={Proceedings of the 46th International ACM SIGIR Conference on Research and Development in Information Retrieval},
  pages={2431--2436},
  year={2023}
}

@article{hernandez2024fraud,
  title={Financial fraud detection through the application of machine learning techniques: a literature review},
  author={Hernandez Aros, Luisa and Bustamante Molano, Ximena and Gutierrez-Portela, Francisco and Moreno Hernandez, Juan Jose},
  journal={Palgrave Communications},
  volume={11},
  number={1},
  pages={1--15},
  year={2024},
  publisher={Palgrave}
}

@article{mashrur2020risk,
  title={Machine learning for financial risk management: A survey},
  author={Mashrur, Akib and Luo, Wei and Zaidi, Nayyar A and Robles-Kelly, Antonio},
  journal={IEEE Access},
  volume={8},
  pages={203203--203223},
  year={2020},
  publisher={IEEE}
}

@article{ozbayoglu2020deep,
  title={Deep learning for financial applications: A survey},
  author={Ozbayoglu, Ahmet Murat and Gudelek, Mehmet Ugur and Sezer, Omer Berat},
  journal={Applied Soft Computing},
  volume={93},
  pages={106384},
  year={2020},
  publisher={Elsevier}
}

@article{hojaji2022behavioral,
  title={Machine Learning in Behavioral Finance: A Systematic Literature Review},
  author={Hojaji, S Navid and Yahyazadehfar, Mahmood and Abedin, Bahareh},
  journal={The Journal of Financial Data Science},
  volume={4},
  number={3},
  pages={129--146},
  year={2022},
  publisher={Institutional Investor Journals}
}

@article{zhao2023mbsrs,
  title={MbSRS: A multi-behavior streaming recommender system},
  author={Zhao, Yan and Wang, Shoujin and Wang, Yan and Liu, Hongwei},
  journal={Information Sciences},
  volume={631},
  pages={1--17},
  year={2023},
  publisher={Elsevier}
}

@article{gayoavello2009session,
  title={A survey on session detection methods in query logs and a proposal for future evaluation},
  author={Gayo-Avello, Daniel},
  journal={Information Sciences},
  volume={179},
  number={12},
  pages={1822--1843},
  year={2009},
  publisher={Elsevier}
}

@inproceedings{covington2016youtube,
  title={Deep neural networks for YouTube recommendations},
  author={Covington, Paul and Adams, Jay and Sargin, Emre},
  booktitle={Proceedings of the 10th ACM Conference on Recommender Systems},
  pages={191--198},
  year={2016}
}

@article{amatriain2013netflix,
  title={Big \& personal: data and models behind Netflix recommendations},
  author={Amatriain, Xavier and Basilico, Justin},
  journal={ACM SIGKDD Explorations Newsletter},
  volume={14},
  number={2},
  pages={37--48},
  year={2013},
  publisher={ACM}
}

@article{lashkari2017survey,
  title={A survey on user profiling model for anomaly detection in cyberspace},
  author={Lashkari, Ahmad H and Chen, Meng and Ghorbani, Ali A},
  journal={Journal of Information Security and Applications},
  volume={34},
  pages={38--56},
  year={2017},
  publisher={Elsevier}
}

@article{wang2017clickstream,
  title={Clickstream user behavior models},
  author={Wang, Gang and Zhang, Xinyang and Tang, Shaomei and Wilson, Christo and Zheng, Haitao and Zhao, Ben Y},
  journal={ACM Transactions on the Web (TWEB)},
  volume={11},
  number={4},
  pages={1--37},
  year={2017},
  publisher={ACM}
}

@article{guo2020survey,
  title={A survey on knowledge graph-based recommender systems},
  author={Guo, Qingyu and Zhuang, Fuzhen and Qin, Chuan and Zhu, Hengshu and Xie, Xing and Xiong, Hui and He, Qing},
  journal={IEEE Transactions on Knowledge and Data Engineering},
  volume={34},
  number={8},
  pages={3549--3568},
  year={2020},
  publisher={IEEE}
}

@article{wu2022graph,
  title={Graph neural networks in recommender systems: a survey},
  author={Wu, Shiwen and Sun, Fei and Zhang, Wentao and Xie, Xu and Cui, Bin},
  journal={ACM Computing Surveys},
  volume={55},
  number={5},
  pages={1--37},
  year={2022},
  publisher={ACM New York, NY}
}

@article{li2024graph,
  title={Graph and sequential neural networks in session-based recommendation: A survey},
  author={Li, Zihao and Yang, Chao and Chen, Yakun and Wang, Xianzhi and Chen, Hongxu and Xu, Guandong and Yao, Lina and Sheng, Michael},
  journal={ACM Computing Surveys},
  volume={57},
  number={2},
  pages={1--37},
  year={2024},
  publisher={ACM New York, NY}
}

@article{Chen2024ASO,
  title={A Survey on Cross-Domain Sequential Recommendation},
  author={Shu Chen and Zitao Xu and Weike Pan and Qiang Yang and Zhong Ming},
  journal={ArXiv},
  year={2024},
  volume={abs/2401.04971},
}

@article{pan2024survey,
  title={A Survey on Sequential Recommendation},
  author={Pan, Liwei and Pan, Weike and Wei, Meiyan and Yin, Hongzhi and Ming, Zhong},
  journal={arXiv preprint arXiv:2412.12770},
  year={2024}
}

@article{boka2024survey,
  title={A survey of sequential recommendation systems: Techniques, evaluation, and future directions},
  author={Boka, Tesfaye Fenta and Niu, Zhendong and Neupane, Rama Bastola},
  journal={Information Systems},
  pages={102427},
  year={2024},
  publisher={Elsevier}
}

@inproceedings{sun2019bert4rec,
  title={BERT4Rec: Sequential recommendation with bidirectional encoder representations from transformer},
  author={Sun, Fei and Liu, Jun and Wu, Jian and Pei, Changhua and Lin, Xiao and Ou, Wenwu and Jiang, Peng},
  booktitle={Proceedings of the 28th ACM international conference on information and knowledge management},
  pages={1441--1450},
  year={2019}
}

@article{Tan2016ImprovedRN,
  title={Improved Recurrent Neural Networks for Session-based Recommendations},
  author={Yong Kiam Tan and Xinxing Xu and Yong Liu},
  journal={Proceedings of the 1st Workshop on Deep Learning for Recommender Systems},
  year={2016},
}

@article{Kang2018SelfAttentiveSR,
  title={Self-Attentive Sequential Recommendation},
  author={Wang-Cheng Kang and Julian McAuley},
  journal={2018 IEEE International Conference on Data Mining (ICDM)},
  year={2018},
  pages={197-206},
}

@article{liguori2023modeling,
  title={Modeling Events and Interactions through Temporal Processes--A Survey},
  author={Liguori, Angelica and Caroprese, Luciano and Minici, Marco and Veloso, Bruno and Spinnato, Francesco and Nanni, Mirco and Manco, Giuseppe and Gama, Joao},
  journal={arXiv preprint arXiv:2303.06067},
  year={2023}
}

@inproceedings{tgn_icml_grl2020,
    title={Temporal Graph Networks for Deep Learning on Dynamic Graphs},
    author={Emanuele Rossi and Ben Chamberlain and Fabrizio Frasca and Davide Eynard and Federico 
    Monti and Michael Bronstein},
    booktitle={ICML 2020 Workshop on Graph Representation Learning},
    year={2020}
}

@article{xu2020inductive,
  title={Inductive representation learning on temporal graphs},
  author={Xu, Da and Ruan, Chuanwei and Korpeoglu, Evren and Kumar, Sushant and Achan, Kannan},
  journal={arXiv preprint arXiv:2002.07962},
  year={2020}
}

@inproceedings{trivedi2019dyrep,
  title={Dyrep: Learning representations over dynamic graphs},
  author={Trivedi, Rakshit and Farajtabar, Mehrdad and Biswal, Prasenjeet and Zha, Hongyuan},
  booktitle={International conference on learning representations},
  year={2019}
}

@inproceedings{nguyen2018continuous,
  title={Continuous-time dynamic network embeddings},
  author={Nguyen, Giang Hoang and Lee, John Boaz and Rossi, Ryan A and Ahmed, Nesreen K and Koh, Eunyee and Kim, Sungchul},
  booktitle={Companion Proceedings of the The Web Conference 2018},
  pages={969--976},
  year={2018}
}

@article{huang2023temporal,
  title={Temporal graph benchmark for machine learning on temporal graphs},
  author={Huang, Shenyang and Poursafaei, Farimah and Danovitch, Jacob and Fey, Matthias and Hu, Weihua and Rossi, Emanuele and Leskovec, Jure and Bronstein, Michael and Rabusseau, Guillaume and Rabbany, Reihaneh},
  journal={Advances in Neural Information Processing Systems},
  volume={36},
  pages={2056--2073},
  year={2023}
}

@article{gastinger2024tgb,
  title={Tgb 2.0: A benchmark for learning on temporal knowledge graphs and heterogeneous graphs},
  author={Gastinger, Julia and Huang, Shenyang and Galkin, Michael and Loghmani, Erfan and Parviz, Ali and Poursafaei, Farimah and Danovitch, Jacob and Rossi, Emanuele and Koutis, Ioannis and Stuckenschmidt, Heiner and others},
  journal={Advances in neural information processing systems},
  volume={37},
  pages={140199--140229},
  year={2024}
}

@article{yi2025tgb,
  title={TGB-Seq Benchmark: Challenging Temporal GNNs with Complex Sequential Dynamics},
  author={Yi, Lu and Peng, Jie and Zheng, Yanping and Mo, Fengran and Wei, Zhewei and Ye, Yuhang and Zixuan, Yue and Huang, Zengfeng},
  journal={arXiv preprint arXiv:2502.02975},
  year={2025}
}

@inproceedings{cho2011friendship,
  title={Friendship and mobility: user movement in location-based social networks},
  author={Cho, Eunjoon and Myers, Seth A and Leskovec, Jure},
  booktitle={Proceedings of the 17th ACM SIGKDD international conference on Knowledge discovery and data mining},
  pages={1082--1090},
  year={2011}
}

@misc{snapnets,
  author       = {Jure Leskovec and Andrej Krevl},
  title        = {{SNAP Datasets}: {Stanford} Large Network Dataset Collection},
  howpublished = {\url{http://snap.stanford.edu/data}},
  month        = jun,
  year         = 2014
}

@article{morton1966computer,
  title={A computer oriented geodetic data base and a new technique in file sequencing},
  author={Morton, Guy M},
  journal={IBM},
  year={1966},
  publisher={IBM}
}

@article{niemeyer2008geohash,
  title={Geohash},
  author={Niemeyer, Gustavo},
  journal={Geohash},
  year={2008}
}

@inproceedings{mcauley2015image,
  title={Image-based recommendations on styles and substitutes},
  author={McAuley, Julian and Targett, Christopher and Shi, Qinfeng and Van Den Hengel, Anton},
  booktitle={Proceedings of the 38th international ACM SIGIR conference on research and development in information retrieval},
  pages={43--52},
  year={2015}
}

@article{behrouz2024best,
  title={Best of both worlds: Advantages of hybrid graph sequence models},
  author={Behrouz, Ali and Parviz, Ali and Karami, Mahdi and Sanford, Clayton and Perozzi, Bryan and Mirrokni, Vahab},
  journal={arXiv preprint arXiv:2411.15671},
  year={2024}
}

@inproceedings{velivckovic2018graph,
  title={Graph Attention Networks},
  author={Veli{\v{c}}kovi{\'c}, Petar and Cucurull, Guillem and Casanova, Arantxa and Romero, Adriana and Li{\`o}, Pietro and Bengio, Yoshua},
  booktitle={International Conference on Learning Representations},
  year={2018}
}

@inproceedings{hu2020heterogeneous,
  title={Heterogeneous graph transformer},
  author={Hu, Ziniu and Dong, Yuxiao and Wang, Kuansan and Sun, Yizhou},
  booktitle={Proceedings of the web conference 2020},
  pages={2704--2710},
  year={2020}
}

@article{zhao2025gigl,
  title={GiGL: Large-Scale Graph Neural Networks at Snapchat},
  author={Zhao, Tong and Liu, Yozen and Kolodner, Matthew and Montemayor, Kyle and Ghazizadeh, Elham and Batra, Ankit and Fan, Zihao and Gao, Xiaobin and Guo, Xuan and Ren, Jiwen and others},
  journal={arXiv e-prints},
  pages={arXiv--2502},
  year={2025}
}

@inproceedings{kipf2017semi,
  title={Semi-Supervised Classification with Graph Convolutional Networks},
  author={Kipf, Thomas N and Welling, Max},
  booktitle={International Conference on Learning Representations},
  year={2017}
}

@inproceedings{ying2018graph,
  title={Graph convolutional neural networks for web-scale recommender systems},
  author={Ying, Rex and He, Ruining and Chen, Kaifeng and Eksombatchai, Pong and Hamilton, William L and Leskovec, Jure},
  booktitle={Proceedings of the 24th ACM SIGKDD international conference on knowledge discovery \& data mining},
  pages={974--983},
  year={2018}
}

@article{hamilton2017inductive,
  title={Inductive representation learning on large graphs},
  author={Hamilton, Will and Ying, Zhitao and Leskovec, Jure},
  journal={Advances in neural information processing systems},
  volume={30},
  year={2017}
}

@article{gao2023survey,
  title={A survey of graph neural networks for recommender systems: Challenges, methods, and directions},
  author={Gao, Chen and Zheng, Yu and Li, Nian and Li, Yinfeng and Qin, Yingrong and Piao, Jinghua and Quan, Yuhan and Chang, Jianxin and Jin, Depeng and He, Xiangnan and others},
  journal={ACM Transactions on Recommender Systems},
  volume={1},
  number={1},
  pages={1--51},
  year={2023},
  publisher={ACM New York, NY, USA}
}

@inproceedings{devlin2019bert,
  title={Bert: Pre-training of deep bidirectional transformers for language understanding},
  author={Devlin, Jacob and Chang, Ming-Wei and Lee, Kenton and Toutanova, Kristina},
  booktitle={Proceedings of the 2019 conference of the North American chapter of the association for computational linguistics: human language technologies, volume 1 (long and short papers)},
  pages={4171--4186},
  year={2019}
}

@inproceedings{grover2016node2vec,
  title={node2vec: Scalable feature learning for networks},
  author={Grover, Aditya and Leskovec, Jure},
  booktitle={Proceedings of the 22nd ACM SIGKDD international conference on Knowledge discovery and data mining},
  pages={855--864},
  year={2016}
}

@inproceedings{mei2017neural,
  title={The Neural Hawkes Process: A Neurally Self-Modulating Multivariate Point Process},
  author={Mei, Hongyuan and Eisner, Jason M.},
  booktitle={Advances in Neural Information Processing Systems},
  volume={30},
  pages={6754--6764},
  year={2017}
}

@inproceedings{zuo2020transformer,
  title={Transformer Hawkes Process},
  author={Zuo, Simiao and Jiang, Haoming and Li, Zitong and Zhao, Tuo and Zha, Hongyuan},
  booktitle={International Conference on Machine Learning},
  pages={11692--11702},
  year={2020}
}

@article{hawkes1971spectra,
  title={Spectra of some self-exciting and mutually exciting point processes},
  author={Hawkes, Alan G},
  journal={Biometrika},
  year={1971}
}

@inproceedings{zhang2020self,
  title={Self-attentive Hawkes process},
  author={Zhang, Qiang and Lipani, Aldo and Kirnap, Omer and Yilmaz, Emine},
  booktitle={International conference on machine learning},
  pages={11183--11193},
  year={2020},
  organization={PMLR}
}

@inproceedings{rendle2010factorizing,
  title={Factorizing personalized markov chains for next-basket recommendation},
  author={Rendle, Steffen and Freudenthaler, Christoph and Schmidt-Thieme, Lars},
  booktitle={Proceedings of the 19th international conference on World wide web},
  pages={811--820},
  year={2010}
}

@article{hidasi2015session,
  title={Session-based recommendations with recurrent neural networks},
  author={Hidasi, Bal{\'a}zs and Karatzoglou, Alexandros and Baltrunas, Linas and Tikk, Domonkos},
  journal={arXiv preprint arXiv:1511.06939},
  year={2015}
}

@inproceedings{kang2018self,
  title={Self-attentive sequential recommendation},
  author={Kang, Wang-Cheng and McAuley, Julian},
  booktitle={2018 IEEE international conference on data mining (ICDM)},
  pages={197--206},
  year={2018},
  organization={IEEE}
}

@article{han2020graph,
  title={Graph hawkes neural network for forecasting on temporal knowledge graphs},
  author={Han, Zhen and Ma, Yunpu and Wang, Yuyi and G{\"u}nnemann, Stephan and Tresp, Volker},
  journal={arXiv preprint arXiv:2003.13432},
  year={2020}
}

@article{kazemi2020representation,
  title={Representation learning for dynamic graphs: A survey},
  author={Kazemi, Seyed Mehran and Goel, Rishab and Jain, Kshitij and Kobyzev, Ivan and Sethi, Akshay and Forsyth, Peter and Poupart, Pascal},
  journal={Journal of Machine Learning Research},
  volume={21},
  number={70},
  pages={1--73},
  year={2020}
}

@article{yin2019dhne,
  title={DHNE: Network representation learning method for dynamic heterogeneous networks},
  author={Yin, Ying and Ji, Li-Xin and Zhang, Jian-Peng and Pei, Yu-Long},
  journal={IEEE Access},
  volume={7},
  pages={134782--134792},
  year={2019},
  publisher={IEEE}
}

@inproceedings{xue2021modeling,
  title={Modeling dynamic heterogeneous network for link prediction using hierarchical attention with temporal rnn},
  author={Xue, Hansheng and Yang, Luwei and Jiang, Wen and Wei, Yi and Hu, Yi and Lin, Yu},
  booktitle={Machine Learning and Knowledge Discovery in Databases: European Conference, ECML PKDD 2020, Ghent, Belgium, September 14--18, 2020, Proceedings, Part I},
  pages={282--298},
  year={2021},
  organization={Springer}
}

@inproceedings{bian2019network,
  title={Network embedding and change modeling in dynamic heterogeneous networks},
  author={Bian, Ranran and Koh, Yun Sing and Dobbie, Gillian and Divoli, Anna},
  booktitle={Proceedings of the 42nd international ACM SIGIR conference on research and development in information retrieval},
  pages={861--864},
  year={2019}
}

@article{hu2020open,
  title={Open graph benchmark: Datasets for machine learning on graphs},
  author={Hu, Weihua and Fey, Matthias and Zitnik, Marinka and Dong, Yuxiao and Ren, Hongyu and Liu, Bowen and Catasta, Michele and Leskovec, Jure},
  journal={Advances in neural information processing systems},
  volume={33},
  pages={22118--22133},
  year={2020}
}

@article{hu2021ogb,
  title={Ogb-lsc: A large-scale challenge for machine learning on graphs},
  author={Hu, Weihua and Fey, Matthias and Ren, Hongyu and Nakata, Maho and Dong, Yuxiao and Leskovec, Jure},
  journal={arXiv preprint arXiv:2103.09430},
  year={2021}
}

@article{yue2022htgn,
  title={HTGN-BTW: Heterogeneous temporal graph network with bi-time-window training strategy for temporal link prediction},
  author={Yue, Chongjian and Du, Lun and Fu, Qiang and Bi, Wendong and Liu, Hengyu and Gu, Yu and Yao, Di},
  journal={arXiv preprint arXiv:2202.12713},
  year={2022}
}

@inproceedings{li2023simplifying,
  title={Simplifying temporal heterogeneous network for continuous-time link prediction},
  author={Li, Ce and Hong, Rongpei and Xu, Xovee and Trajcevski, Goce and Zhou, Fan},
  booktitle={Proceedings of the 32nd ACM International Conference on Information and Knowledge Management},
  pages={1288--1297},
  year={2023}
}

@inproceedings{wu2020mind,
  title={Mind: A large-scale dataset for news recommendation},
  author={Wu, Fangzhao and Qiao, Ying and Chen, Jiun-Hung and Wu, Chuhan and Qi, Tao and Lian, Jianxun and Liu, Danyang and Xie, Xing and Gao, Jianfeng and Wu, Winnie and others},
  booktitle={Proceedings of the 58th annual meeting of the association for computational linguistics},
  pages={3597--3606},
  year={2020}
}

@article{yuan2022tenrec,
  title={Tenrec: A large-scale multipurpose benchmark dataset for recommender systems},
  author={Yuan, Guanghu and Yuan, Fajie and Li, Yudong and Kong, Beibei and Li, Shujie and Chen, Lei and Yang, Min and Yu, Chenyun and Hu, Bo and Li, Zang and others},
  journal={Advances in Neural Information Processing Systems},
  volume={35},
  pages={11480--11493},
  year={2022}
}

@article{zhang2024ninerec,
  title={Ninerec: A benchmark dataset suite for evaluating transferable recommendation},
  author={Zhang, Jiaqi and Cheng, Yu and Ni, Yongxin and Pan, Yunzhu and Yuan, Zheng and Fu, Junchen and Li, Youhua and Wang, Jie and Yuan, Fajie},
  journal={IEEE Transactions on Pattern Analysis and Machine Intelligence},
  year={2024},
  publisher={IEEE}
}

@inproceedings{zhu2022bars,
  title={Bars: Towards open benchmarking for recommender systems},
  author={Zhu, Jieming and Dai, Quanyu and Su, Liangcai and Ma, Rong and Liu, Jinyang and Cai, Guohao and Xiao, Xi and Zhang, Rui},
  booktitle={Proceedings of the 45th International ACM SIGIR Conference on Research and Development in Information Retrieval},
  pages={2912--2923},
  year={2022}
}

@article{osin2024ebes,
  title={EBES: Easy Benchmarking for Event Sequences},
  author={Osin, Dmitry and Udovichenko, Igor and Moskvoretskii, Viktor and Shvetsov, Egor and Burnaev, Evgeny},
  journal={arXiv preprint arXiv:2410.03399},
  year={2024}
}

@article{xue2023easytpp,
  title={Easytpp: Towards open benchmarking temporal point processes},
  author={Xue, Siqiao and Shi, Xiaoming and Chu, Zhixuan and Wang, Yan and Hao, Hongyan and Zhou, Fan and Jiang, Caigao and Pan, Chen and Zhang, James Y and Wen, Qingsong and others},
  journal={arXiv preprint arXiv:2307.08097},
  year={2023}
}

@article{karpukhin2024hotpp,
  title={HoTPP Benchmark: Are We Good at the Long Horizon Events Forecasting?},
  author={Karpukhin, Ivan and Shipilov, Foma and Savchenko, Andrey},
  journal={arXiv preprint arXiv:2406.14341},
  year={2024}
}

@inproceedings{li2021temporal,
  title={Temporal knowledge graph reasoning based on evolutional representation learning},
  author={Li, Zixuan and Jin, Xiaolong and Li, Wei and Guan, Saiping and Guo, Jiafeng and Shen, Huawei and Wang, Yuanzhuo and Cheng, Xueqi},
  booktitle={Proceedings of the 44th international ACM SIGIR conference on research and development in information retrieval},
  pages={408--417},
  year={2021}
}

@article{jin2019recurrent,
  title={Recurrent event network: Autoregressive structure inference over temporal knowledge graphs},
  author={Jin, Woojeong and Qu, Meng and Jin, Xisen and Ren, Xiang},
  journal={arXiv preprint arXiv:1904.05530},
  year={2019}
}

@article{li2022complex,
  title={Complex evolutional pattern learning for temporal knowledge graph reasoning},
  author={Li, Zixuan and Guan, Saiping and Jin, Xiaolong and Peng, Weihua and Lyu, Yajuan and Zhu, Yong and Bai, Long and Li, Wei and Guo, Jiafeng and Cheng, Xueqi},
  journal={arXiv preprint arXiv:2203.07782},
  year={2022}
}

@article{wang2024graph,
  title={Graph-mamba: Towards long-range graph sequence modeling with selective state spaces},
  author={Wang, Chloe and Tsepa, Oleksii and Ma, Jun and Wang, Bo},
  journal={arXiv preprint arXiv:2402.00789},
  year={2024}
}

@inproceedings{behrouz2024graph,
  title={Graph mamba: Towards learning on graphs with state space models},
  author={Behrouz, Ali and Hashemi, Farnoosh},
  booktitle={Proceedings of the 30th ACM SIGKDD conference on knowledge discovery and data mining},
  pages={119--130},
  year={2024}
}

@article{huang2024can,
  title={What Can We Learn from State Space Models for Machine Learning on Graphs?},
  author={Huang, Yinan and Miao, Siqi and Li, Pan},
  journal={arXiv preprint arXiv:2406.05815},
  year={2024}
}

@article{ding2024dygmamba,
  title={DyGMamba: Efficiently Modeling Long-Term Temporal Dependency on Continuous-Time Dynamic Graphs with State Space Models},
  author={Ding, Zifeng and Li, Yifeng and He, Yuan and Norelli, Antonio and Wu, Jingcheng and Tresp, Volker and Ma, Yunpu and Bronstein, Michael},
  journal={arXiv preprint arXiv:2408.04713},
  year={2024}
}

@article{li2024dyg,
  title={Dyg-mamba: Continuous state space modeling on dynamic graphs},
  author={Li, Dongyuan and Tan, Shiyin and Zhang, Ying and Jin, Ming and Pan, Shirui and Okumura, Manabu and Jiang, Renhe},
  journal={arXiv preprint arXiv:2408.06966},
  year={2024}
}

@article{hai2024study,
  title={A Study of Recommendation Methods Based on Graph Hybrid Neural Networks and Deep Crossing},
  author={Hai, Yan and Wang, Dongyang and Liu, Zhizhong and Zheng, Jitao and Ding, Chengrui},
  journal={Electronics},
  volume={13},
  number={21},
  pages={4224},
  year={2024},
  publisher={MDPI}
}

@inproceedings{shi2019two,
  title={Two-stream adaptive graph convolutional networks for skeleton-based action recognition},
  author={Shi, Lei and Zhang, Yifan and Cheng, Jian and Lu, Hanqing},
  booktitle={Proceedings of the IEEE/CVF conference on computer vision and pattern recognition},
  pages={12026--12035},
  year={2019}
}

@article{yu2017spatio,
  title={Spatio-temporal graph convolutional networks: A deep learning framework for traffic forecasting},
  author={Yu, Bing and Yin, Haoteng and Zhu, Zhanxing},
  journal={arXiv preprint arXiv:1709.04875},
  year={2017}
}

@inproceedings{yu2020spatio,
  title={Spatio-temporal graph transformer networks for pedestrian trajectory prediction},
  author={Yu, Cunjun and Ma, Xiao and Ren, Jiawei and Zhao, Haiyu and Yi, Shuai},
  booktitle={Computer Vision--ECCV 2020: 16th European Conference, Glasgow, UK, August 23--28, 2020, Proceedings, Part XII 16},
  pages={507--523},
  year={2020},
  organization={Springer}
}

@article{bai2020adaptive,
  title={Adaptive graph convolutional recurrent network for traffic forecasting},
  author={Bai, Lei and Yao, Lina and Li, Can and Wang, Xianzhi and Wang, Can},
  journal={Advances in neural information processing systems},
  volume={33},
  pages={17804--17815},
  year={2020}
}

@inproceedings{wu2020connecting,
  title={Connecting the dots: Multivariate time series forecasting with graph neural networks},
  author={Wu, Zonghan and Pan, Shirui and Long, Guodong and Jiang, Jing and Chang, Xiaojun and Zhang, Chengqi},
  booktitle={Proceedings of the 26th ACM SIGKDD international conference on knowledge discovery \& data mining},
  pages={753--763},
  year={2020}
}

@article{zhang2001graph,
  title={Graph-bert: Only attention is needed for learning graph representations. arXiv 2020},
  author={Zhang, Jiawei and Zhang, Haopeng and Xia, Congying and Sun, Li},
  journal={arXiv preprint arXiv:2001.05140},
  year={2001}
}

@article{mao2021molecular,
  title={Molecular graph enhanced transformer for retrosynthesis prediction},
  author={Mao, Kelong and Xiao, Xi and Xu, Tingyang and Rong, Yu and Huang, Junzhou and Zhao, Peilin},
  journal={Neurocomputing},
  volume={457},
  pages={193--202},
  year={2021},
  publisher={Elsevier}
}

@inproceedings{liu2020k,
  title={K-bert: Enabling language representation with knowledge graph},
  author={Liu, Weijie and Zhou, Peng and Zhao, Zhe and Wang, Zhiruo and Ju, Qi and Deng, Haotang and Wang, Ping},
  booktitle={Proceedings of the AAAI conference on artificial intelligence},
  volume={34},
  number={03},
  pages={2901--2908},
  year={2020}
}

@inproceedings{yasunaga2021qa,
  title={QA-GNN: Reasoning with Language Models and Knowledge Graphs for Question Answering},
  author={Yasunaga, Michihiro and Ren, Hongyu and Bosselut, Antoine and Liang, Percy and Leskovec, Jure},
  booktitle={North American Chapter of the Association for Computational Linguistics (NAACL)},
  year={2021}
}

@inproceedings{huang2020knowledge,
  title={Knowledge Graph-Augmented Abstractive Summarization with Semantic-Driven Cloze Reward},
  author={Huang, Luyang and Wu, Lingfei and Wang, Lu},
  booktitle={Proceedings of the 58th Annual Meeting of the Association for Computational Linguistics},
  pages={5094--5107},
  year={2020}
}

@article{peng2024graph,
  title={Graph retrieval-augmented generation: A survey},
  author={Peng, Boci and Zhu, Yun and Liu, Yongchao and Bo, Xiaohe and Shi, Haizhou and Hong, Chuntao and Zhang, Yan and Tang, Siliang},
  journal={arXiv preprint arXiv:2408.08921},
  year={2024}
}

@article{zhu2025graph,
  title={Graph-based Approaches and Functionalities in Retrieval-Augmented Generation: A Comprehensive Survey},
  author={Zhu, Zulun and Huang, Tiancheng and Wang, Kai and Ye, Junda and Chen, Xinghe and Luo, Siqiang},
  journal={arXiv preprint arXiv:2504.10499},
  year={2025}
}

@article{letham2013sequential,
  title={Sequential event prediction},
  author={Letham, Benjamin and Rudin, Cynthia and Madigan, David},
  journal={Machine learning},
  volume={93},
  number={2},
  pages={357--380},
  year={2013},
  publisher={Springer}
}

@article{boyd2020user,
  title={User-dependent neural sequence models for continuous-time event data},
  author={Boyd, Alex and Bamler, Robert and Mandt, Stephan and Smyth, Padhraic},
  journal={Advances in Neural Information Processing Systems},
  volume={33},
  pages={21488--21499},
  year={2020}
}

@article{yang2022towards,
  title={Towards out-of-distribution sequential event prediction: A causal treatment},
  author={Yang, Chenxiao and Wu, Qitian and Wen, Qingsong and Zhou, Zhiqiang and Sun, Liang and Yan, Junchi},
  journal={Advances in neural information processing systems},
  volume={35},
  pages={22656--22670},
  year={2022}
}

@article{fey2019fast,
  title={Fast graph representation learning with PyTorch Geometric},
  author={Fey, Matthias and Lenssen, Jan Eric},
  journal={arXiv preprint arXiv:1903.02428},
  year={2019}
}

@article{yu2023towards,
  title={Towards better dynamic graph learning: New architecture and unified library},
  author={Yu, Le and Sun, Leilei and Du, Bowen and Lv, Weifeng},
  journal={Advances in Neural Information Processing Systems},
  volume={36},
  pages={67686--67700},
  year={2023}
}

@inproceedings{luo2022neighborhood,
  title={Neighborhood-aware scalable temporal network representation learning},
  author={Luo, Yuhong and Li, Pan},
  booktitle={Learning on Graphs Conference},
  pages={1--1},
  year={2022},
  organization={PMLR}
}

@article{zhang2406efficient,
  title={Efficient neural common neighbor for temporal graph link prediction, 2024},
  author={Zhang, Xiaohui and Wang, Yanbo and Wang, Xiyuan and Zhang, Muhan},
  journal={URL https://arxiv. org/abs/2406.07926}
}

@article{gravina2024long,
  title={Long range propagation on continuous-time dynamic graphs},
  author={Gravina, Alessio and Lovisotto, Giulio and Gallicchio, Claudio and Bacciu, Davide and Grohnfeldt, Claas},
  journal={arXiv preprint arXiv:2406.02740},
  year={2024}
}

@inproceedings{shi2021masked,
  title={Masked Label Prediction: Unified Message Passing Model for Semi-Supervised Classification},
  author={Shi, Yunsheng and Huang, Zhengjie and Feng, Shikun and Zhong, Hui and Wang, Wenjing and Sun, Yu},
  booktitle={Proceedings of the Thirtieth International Joint Conference on Artificial Intelligence},
  pages={1548--1554},
  year={2021},
  organization={International Joint Conferences on Artificial Intelligence Organization}
}

@article{lu2024improving,
  title={Improving temporal link prediction via temporal walk matrix projection},
  author={Lu, Xiaodong and Sun, Leilei and Zhu, Tongyu and Lv, Weifeng},
  journal={Advances in Neural Information Processing Systems},
  volume={37},
  pages={141153--141182},
  year={2024}
}

@article{gao2025hyperevent,
  title={HyperEvent: Learning Cohesive Events for Large-scale Dynamic Link Prediction},
  author={Gao, Jian and Wu, Jianshe and Ding, JingYi},
  journal={arXiv preprint arXiv:2507.11836},
  year={2025}
}

\newpage

\appendix

\section{Dataset Documentation}
\label{sec:ap-dataset}

All datasets presented in this paper are intended for academic research purposes, and their corresponding licenses are listed in this section. They are constructed from publicly available resources described below. In all cases, we perform anonymization by removing any personally identifiable information. User IDs in the original data are replaced with auto-incremented ID numbers.

\paragraph{Download links.} 
The datasets and tasks described in this paper are available for download from the following links:

\begin{itemize}
    \item Datasets and prediction tasks website and documentation: \href{https://huggingface.co/datasets/capitalone/PRES}{https://huggingface.co/datasets/
    capitalone/PRES}
    \item Code for dataset creation and experiment runs: \href{https://github.com/CapitalOne-Research/personal-relational-event-sequence}{https://github.com/CapitalOne-Research/
    personal-relational-event-sequence}
\end{itemize}

\paragraph{Dataset source and license information.} Below, we describe how the source data was obtained and provide license information for each dataset:
\begin{itemize}
    \item \underline{\textsc{pres-github}}. This dataset is based on GitHub data collected from the \href{https://www.gharchive.org/}{GH Archive website} (\url{https://www.gharchive.org/}) using its HTTP JSON download link. It contains GitHub user activity from January 2025, and user IDs have been anonymized. Content from GH Archive is released under the \underline{CC-BY-4.0 license}, while the associated code is released under the MIT license.

    \item \underline{\textsc{pres-amazon-clothing}} and \underline{\textsc{pres-amazon-electronics}}. These datasets contain Amazon product reviews and ratings in their respective categories. Both are based on Amazon review data collected by \citet{mcauley2015image} and hosted at: \url{https://cseweb.ucsd.edu/~jmcauley/datasets/amazon/links.html}. The Amazon review content is licensed under the \underline{Amazon license}:
    \begin{quote}
         By accessing the Amazon Customer Reviews Library ("Reviews Library"), you agree that the Reviews Library is an Amazon Service subject to the \href{https://www.amazon.com/gp/help/customer/display.html?nodeId=508088}{Amazon.com Conditions of Use} and you agree to be bound by them, with the following additional conditions:
    
    In addition to the license rights granted under the Conditions of Use, Amazon or its content providers grant you a limited, non-exclusive, non-transferable, non-sublicensable, revocable license to access and use the Reviews Library for purposes of academic research. You may not resell, republish, or make any commercial use of the Reviews Library or its contents, including use of the Reviews Library for commercial research, such as research related to a funding or consultancy contract, internship, or other relationship in which the results are provided for a fee or delivered to a for-profit organization. You may not (a) link or associate content in the Reviews Library with any personal information (including Amazon customer accounts), or (b) attempt to determine the identity of the author of any content in the Reviews Library. If you violate any of the foregoing conditions, your license to access and use the Reviews Library will automatically terminate without prejudice to any of the other rights or remedies Amazon may have.        
    \end{quote}

    \item \underline{\textsc{pres-gowalla}}. This dataset contains user activity on the (now defunct) social network Gowalla. It was originally collected by \citet{cho2011friendship} using the platform's public API and published in the \href{https://snap.stanford.edu/data/loc-Gowalla.html}{SNAP Dataset Repository} \cite{snapnets} (\url{https://snap.stanford.edu/data/loc-Gowalla.html}). The curator confirmed that SNAP datasets are free to use, but no specific license information is available.

    \item \underline{\textsc{pres-brightkite}}. This dataset contains user activity on the (also now defunct) social network Brightkite. It was also originally collected by \citet{cho2011friendship} using the platform's public API and published in the \href{https://snap.stanford.edu/data/loc-brightkite.html}{SNAP Dataset Repository} \cite{snapnets} (\url{https://snap.stanford.edu/data/loc-brightkite.html}). The curator confirmed that SNAP datasets are free to use, but no specific license information is available.

\end{itemize}


\section{Dataset Contents}
\label{sec:ap-dataset-stats}

\paragraph{Examples of dataset contents.} To illustrate the structure of the curated datasets, we provide examples of user event sequences from several \textsc{pres} datasets. Each table includes both personal and relational events, showing how different types of user activity are represented in our format.

\begin{itemize}
    \item \underline{\textsc{pres-brightkite}} and \underline{\textsc{pres-gowalla}}

\begin{table}[h]
\centering
\footnotesize
\caption{Example of user event sequence in \textsc{pres-brightkite} and \textsc{pres-gowalla}.}
\label{tab:example-events}
\ttfamily
\begin{tabular}{ccccc}
\toprule
\texttt{uid} & \texttt{timestamp} & \texttt{event\_set} & \texttt{event} & \texttt{other\_uid} \\
\midrule
\texttt{39} & \texttt{1206596784} & \texttt{personal}   & \texttt{9xj6hwkm}  & \texttt{<NA>} \\
\texttt{39} & \texttt{1206596838} & \texttt{personal}   & \texttt{9xj3fynm}  & \texttt{<NA>} \\
\texttt{39} & \texttt{1206596871} & \texttt{personal}   & \texttt{9xj3fynm}  & \texttt{<NA>} \\
\texttt{39} & \texttt{1235862855} & \texttt{personal}   & \texttt{9xj65423}  & \texttt{<NA>} \\
\texttt{39} & \texttt{1250883230} & \texttt{personal}   & \texttt{9xj65423}  & \texttt{<NA>} \\
\texttt{39} & \texttt{1254535157} & \texttt{personal}   & \texttt{9xj5skbn}  & \texttt{<NA>} \\
\texttt{39} & \texttt{1254535193} & \texttt{personal}   & \texttt{9xj5sm00}  & \texttt{<NA>} \\
\texttt{39} & \texttt{1283443369} & \texttt{personal}   & \texttt{9q8yyyhs}  & \texttt{<NA>} \\
\texttt{39} & \texttt{<NA>}       & \texttt{relational} & \texttt{friendship} & \texttt{0} \\
\texttt{39} & \texttt{<NA>}       & \texttt{relational} & \texttt{friendship} & \texttt{30} \\
\texttt{39} & \texttt{<NA>}       & \texttt{relational} & \texttt{friendship} & \texttt{105} \\
\texttt{39} & \texttt{<NA>}       & \texttt{relational} & \texttt{friendship} & \texttt{1190} \\
\bottomrule
\end{tabular}
\rmfamily
\end{table}

\item \underline{\textsc{pres-amazon-clothing}} and \underline{\textsc{pres-amazon-electronics}}

\begin{table}[h]
\centering
\footnotesize
\caption{Example of user event sequence in \textsc{pres-amazon-clothing} and \textsc{pres-amazon-electronics}.}
\label{tab:amazon-example}
\ttfamily
\begin{tabular}{ccccc}
\toprule
\texttt{uid} & \texttt{timestamp} & \texttt{event\_set} & \texttt{event} & \texttt{other\_uid} \\
\midrule
\texttt{254057} & \texttt{1375401600} & \texttt{personal}   & \texttt{B000A6PPOK:3}       & \texttt{<NA>} \\
\texttt{254057} & \texttt{1377302400} & \texttt{personal}   & \texttt{B003TMPHOU:5}       & \texttt{<NA>} \\
\texttt{254057} & \texttt{1377302400} & \texttt{personal}   & \texttt{B004A81PJI:4}       & \texttt{<NA>} \\
\texttt{254057} & \texttt{1377302400} & \texttt{personal}   & \texttt{B0054R4AXW:5}       & \texttt{<NA>} \\
\texttt{254057} & \texttt{1377302400} & \texttt{personal}   & \texttt{B005CPGHAA:5}       & \texttt{<NA>} \\
\texttt{254057} & \texttt{1377302400} & \texttt{personal}   & \texttt{B007FNXMEQ:5}       & \texttt{<NA>} \\
\texttt{254057} & \texttt{1377302400} & \texttt{personal}   & \texttt{B007IV7KRU:5}       & \texttt{<NA>} \\
\texttt{254057} & \texttt{1377302400} & \texttt{personal}   & \texttt{B007WAWHD4:5}       & \texttt{<NA>} \\
\texttt{254057} & \texttt{1377302400} & \texttt{personal}   & \texttt{B008AST7R6:5}       & \texttt{<NA>} \\
\texttt{254057} & \texttt{1377302400} & \texttt{personal}   & \texttt{B008R56H4S:5}       & \texttt{<NA>} \\
\texttt{254057} & \texttt{1404086400} & \texttt{relational} & \texttt{co-review\_product} & \texttt{107741} \\
\bottomrule
\end{tabular}
\rmfamily
\end{table}

\item \underline{\textsc{pres-github}}

\begin{table}[h]
\centering
\footnotesize
\caption{Example of user event sequence in \textsc{pres-github}.}
\label{tab:github-example}
\ttfamily
\begin{tabular}{ccccc}
\toprule
\texttt{uid} & \texttt{timestamp} & \texttt{event\_set} & \texttt{event} & \texttt{other\_uid} \\
\midrule
\texttt{3669059} & \texttt{1738288160} & \texttt{personal}   & \texttt{PullRequestReviewCreated} & \texttt{<NA>} \\
\texttt{3669059} & \texttt{1738288191} & \texttt{personal}   & \texttt{PullRequestReviewCreated} & \texttt{<NA>} \\
\texttt{3669059} & \texttt{1738288198} & \texttt{personal}   & \texttt{PullRequestClosed}        & \texttt{<NA>} \\
\texttt{3669059} & \texttt{1738288200} & \texttt{personal}   & \texttt{Push}                     & \texttt{<NA>} \\
\texttt{3669059} & \texttt{1738288206} & \texttt{personal}   & \texttt{PullRequestClosed}        & \texttt{<NA>} \\
\texttt{3669059} & \texttt{1738288207} & \texttt{personal}   & \texttt{Push}                     & \texttt{<NA>} \\
\texttt{3669059} & \texttt{1738288217} & \texttt{personal}   & \texttt{DeleteBranch}             & \texttt{<NA>} \\
\texttt{3669059} & \texttt{1738288219} & \texttt{personal}   & \texttt{DeleteBranch}             & \texttt{<NA>} \\
\texttt{3669059} & \texttt{<NA>}       & \texttt{relational} & \texttt{collaborate}              & \texttt{824409} \\
\texttt{3669059} & \texttt{<NA>}       & \texttt{relational} & \texttt{collaborate}              & \texttt{3126262} \\
\bottomrule
\end{tabular}
\rmfamily
\end{table}

\end{itemize}

\begin{figure}[ht]
\centering

\begin{subfigure}{0.48\textwidth}
    \includegraphics[width=\linewidth]{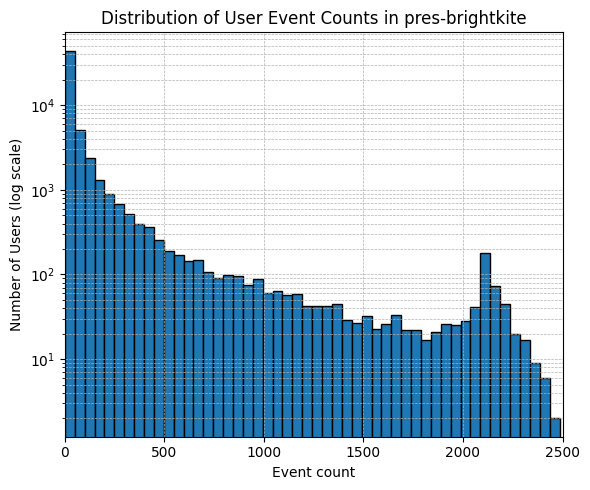}
    \caption{\textsc{pres-brightkite}}
\end{subfigure}
\hfill
\begin{subfigure}{0.48\textwidth}
    \includegraphics[width=\linewidth]{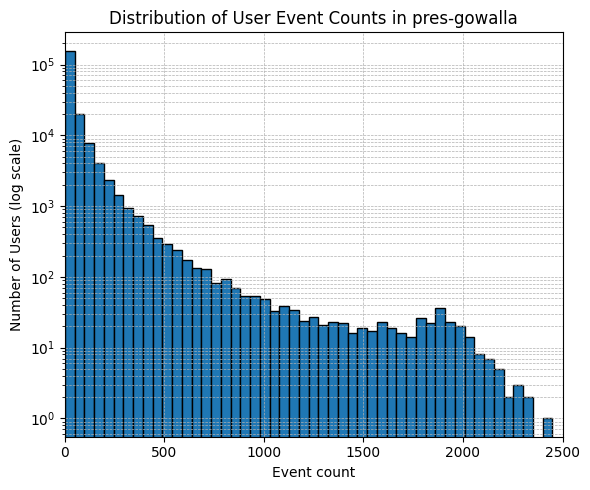}
    \caption{\textsc{pres-gowalla}}
\end{subfigure}

\vspace{1em}  

\begin{subfigure}{0.48\textwidth}
    \includegraphics[width=\linewidth]{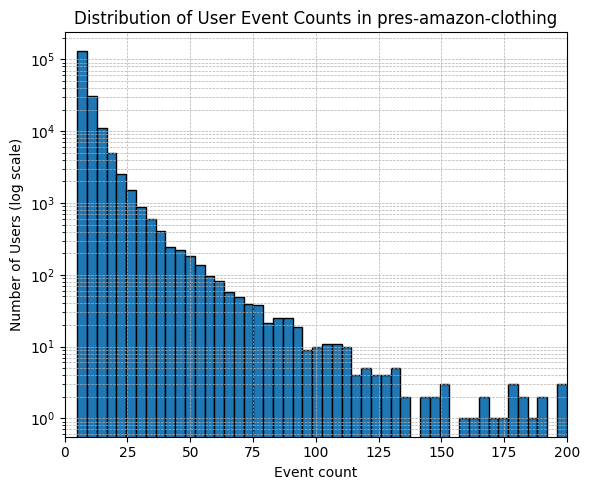}
    \caption{\textsc{pres-amazon-clothing}}
\end{subfigure}
\hfill
\begin{subfigure}{0.48\textwidth}
    \includegraphics[width=\linewidth]{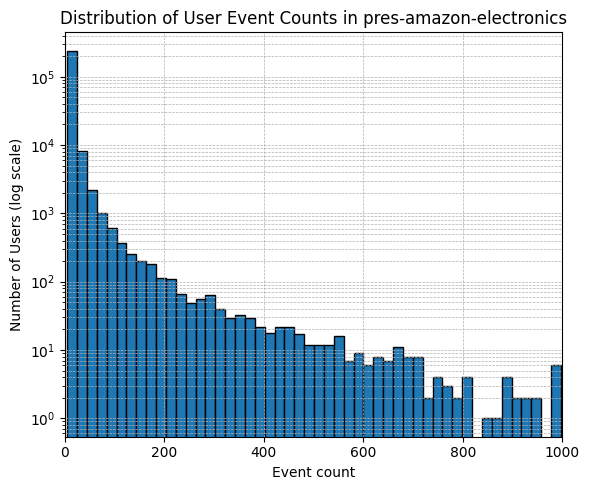}
    \caption{\textsc{pres-amazon-electronics}}
\end{subfigure}

\caption{Histogram of the number of events per user in each dataset.}
\label{fig:event-count}
\end{figure}

\begin{wrapfigure}[16]{r}{0.5\textwidth}
  \centering
  \includegraphics[width=0.5\textwidth]{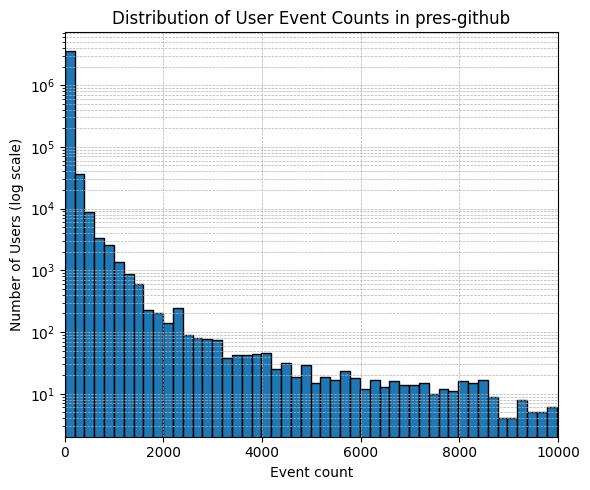}
  \caption{Histogram of the number of event per user in \textsc{pres-github}.}
  \label{fig:event-github}
\end{wrapfigure}
\paragraph{Event statistics.}
To characterize user events, we include histograms in Figure~\ref{fig:event-count} and Figure~\ref{fig:event-github} showing the distribution of event counts per user in each dataset. These histograms are constructed by computing the number of events associated with each user and aggregating how many users fall into each count bucket. The y-axis is log-scaled to highlight the long-tailed nature of user behavior, where the majority of users generate only a small number of events, while a much smaller group contributes disproportionately large volumes of activity. This skew is common across datasets and presents both challenges and opportunities for modeling.


\section{Dataset Construction}
\label{sec:ap-dataset-construction}

This section provides more details about how all of the PRES datasets are created. In short, the data processing files are available in our GitHub repo, following the format \texttt{create\_datasets/process\_X.ipynb} for first generating the processed files, and \texttt{create\_datasets/task\_X.py} for subsequently generating the prediction tasks (including negative sampling for test set), where X is the dataset name.

\subsection{\textsc{pres-brightkite} and \textsc{pres-gowalla}}

The raw data of users’ friendships and check-in histories are stored in text files. We read the friendship files into a dataframe and reformat them into our standard format. 
The check-in history file contains the latitude and longitude of check-in locations. After reading the files, we convert the latitude-longitude encoding of locations into Geohash-8 string representations and convert the check-in times into Unix timestamp format.
Relational events denote friendship between two users, and there are no timestamps associated with this friendship from the raw data. 
We combine both event types, 
sort the data by uid and timestamp, and save it into a CSV file (in our dataset repository on Kaggle, this is \texttt{processed/X\_all\_events.csv}).

Now that we have all events saved in a standardized format, we can create our splits and pre-generate negative samples for reproducibility within \texttt{create\_datasets/task\_brightkite.py} and \texttt{create\_datasets/task\_gowalla.py}. For both tasks, we begin by splitting all events into separate dataframes for personal and relational events.

\paragraph{Relational Task: Friendship Prediction}
To generate the relational event prediction task, we perform a random split of 70\%/10\%/20\% on the relational event dataframe to generate the train/eval/test datasets, and store them into CSV files. Following \citet{gastinger2024tgb}, we adopt a \textit{1-vs-1000} negative sampling scheme, in which 1,000 negative events are sampled for each relational event in the prediction set. Negative samples are drawn via uniform random sampling of users, excluding those who already have relational events with the target user in the training set. 

\paragraph{Personal Task: Check-in Prediction}
To generate the personal event prediction task, we perform timestamp-based splitting on the personal event dataframe into train/val/test sets. We split each user’s personal events by taking the last 20\% for the test set, the previous 10\% for validation, and the remaining 70\% for training. We cap the number of events in the test and validation sets to at most 20 and 10 per user, respectively. Because personal events are more frequent than relational ones, we adopt a \textit{1-vs-500} negative sampling scheme. As geohash strings encode hierarchical spatial information (e.g., earlier characters represent broader regions), we apply stratified hierarchical sampling. Specifically, negatives are stratified by shared geohash prefixes, from matching the first five characters to none. Varying the number of matching characters ensures our negatives contain a mix of nearby and distant locations. 

\subsection{\textsc{pres-amazon-electronics} and \textsc{pres-amazon-clothing}}

The input zipped JSON containing the raw data is loaded into a Pandas DataFrame. Each row corresponds to a single review by a single user. We drop rows belonging to users with fewer than 5 product reviews, and create anonymous user IDs from the input 'reviewID' column.

Personal events here are simply all the remaining rows; the 'event' column is created by concatenating the product ID with the rating given to it in that review. Relational events are created by finding users who have co-reviewed at least 3 of the same products. Timestamps for personal events are naturally the time at which the review was posted. For relational events, the timestamp corresponds to when this co-review condition is first met.

The schema for both of these personal and event dataframes are homogenized. At the end of \texttt{process\_amazon-X.ipynb}, we combine both the relational and personal event data and write it out as a single CSV. This is now the input into \texttt{task\_amazon.py} for each dataset. 

Both tasks use a 70\% train, 10\% validation and 20\% test split paradigm, where the test events are the latest 20\% of events (either personal or relational) per user, with the validation events immediately preceding the test events. To manage large histories of some users, we cap test and val sets at 20 and 10 events per user, respectively.

\paragraph{Relational Task: Co-Review Prediction}
We apply the above logic to only the relational events, and can now identify the maximum timestamp (aka cut-off time) per user in the training data. This is used to split all personal events into 'observed' (before cut-off time) and 'unobserved' (after cut-off time). Only observed personal events are used for training.
Mirroring the logic for \textsc{pres-brightkite} and \textsc{pres-gowalla}'s relational task, we adopt a \textit{1-vs-1000} negative sampling scheme. Each negative corresponds to a uniform random sampling of user IDs that do not have a true co-review relationship with the user ID of interest.

\paragraph{Personal Task: Product Prediction}
Similarly to the relational task, we apply the splitting logic only to personal events. We use the timestamp of the latest train event per user to identify which of that user's relational events are included for training.
We adopt a \textit{1-vs-500} negative sampling scheme, aligning with \textsc{pres-brightkite} and \textsc{pres-gowalla}'s personal task logic. Negative samples for each personal event (e.g., {B001OE3F08:3}) are drawn from three sources: (1) the same product with different ratings (e.g., 
{B001OE3F08:5}); (2) other personal events not in the user’s training data; and (3) samples from the second set with randomly perturbed ratings.





\subsection{\textsc{pres-github}}

The raw data of GitHub users' activities comes as a compressed JSON. We unpack this into a CSV in \texttt{create\_datasets/github\_extract.py}. We aim to remove bot accounts that could add substantial noise by dropping any rows where the user contains \textit{[bot]} in their name or if that user conducted 100,000 or more actions in the dataset.

Next, in \texttt{create\_datasets/process\_github.ipynb} we additionally remove users with fewer than 3 actions. Relational events connect users who have pushed or opened pull requests at least 5 times in the same repository; predicting relational events in this context is about predicting collaborators. Personal events correspond to the rows in the original dataset, which represent only 24 unique GitHub activities. Thus, the personal event task construction used in the previous datasets is not applicable to \textsc{pres-github}. We decided to omit this dataset from the set of datasets used for creating personal event prediction tasks. These personal and relational events have their schemas homogenized and are concatenated. The result is saved as \texttt{processed/github\_all\_events.csv} on Kaggle datasets.

Finally, \texttt{create\_datasets/task\_github.py} is used to create train/validation/test splits and negatives for the relational event prediction task. The logic here follows \textsc{pres-brightkite}, including the split proportions and limitations on the maximum number of events withheld for validation and test. The notable difference is that we adopt a \textit{1-vs-300} negative sampling scheme due to the large size of the dataset.

\section{Additional Related Works}

\label{sec:add_rel_work}


\subsection{Event sequence.}
Event sequence modeling is a broad topic that covers many different domains which share a similar goal of understanding and potentially predicting future events given past history. In healthcare, being able to predicting patient's upcoming medical visits enables proactive care and better resource allocation for healthcare providers. In manufacturing and industrial setting, modeling sequences of equipment sensor readings or machine states allow for early detection of faults, enabling predictive maintenance and reducing downtime. These problems share common challenges such as modeling temporal dependencies, handling irregular and asynchronous observations.

Temporal point processes (TPPs) provide a powerful tool for modeling discrete events as stochastic processes in continuous time. Classical models like the Poison process and the Hawkes process \citep{hawkes1971spectra} allow for explicit modeling of event dynamics, including self-excitation and mutual inhibition, through parameterized intensity functions that govern the likelihood of future events. Recent work has introduced neural extensions such as Neural Hawkes Process \citep{mei2017neural} \citep{zuo2020transformer} and the Self-Attentive Hawkes Process \citep{zhang2020self}, which integrate RNNs or attention mechanisms into the intensity function. These models are particularly well suited for fine-grained timestamp prediction and have found applications in finance, healthcare, and user behavior modeling. However, TPPs typically assume a simple structure over events and focus solely on the temporal dimension, making them less suitable for capturing structured dependencies across users or networks.


\paragraph{Sequential recommendation.} 
A closely related application domain is sequential recommendation, where the goal is to predict the next item a user will interact with based on their historical behavior. Early methods employed Markov chains or matrix factorization over time-slided data \citep{rendle2010factorizing}, 
while more recent models use deep sequence encoders such as GRU4Rec \citep{hidasi2015session}, SASRec \citep{kang2018self}, and BERT4Rec \citep{sun2019bert4rec}, which apply Transformer-based architectures to item sequences. These models have shown strong performance by capturing user preferences over time. However, they typically model each user independently and do not account for interactions among users. 


\subsection{Graph models.} 
In parallel, there has been significant progress in graph-based modeling of user interaction, especially through Graph Neural Networks (GNN). While static graphs lack explicit timestamps and do not capture the order of interactions, they can still represent temporal data through careful graph construction, such as building graphs over specific time windows or pruning outdated edges \citep{zhao2025gigl}. This setup allows for framing personal and relational event prediction tasks as link prediction (e.g. forecasting probability of user-item interactions) or node classification (e.g. fraud detection). Early GNNs like GCN \citep{kipf2017semi} introduced neighborhood aggregation but were limited by their transductive nature and the requirement of full graph knowledge during training. GraphSAGE \citep{hamilton2017inductive} addressed these issues by introducing inductive learning via stochastic sampling, while PinSAGE \citep{ying2018graph} scaled GNNs to billions of nodes via relaxing memory constraints. GAT \citep{velivckovic2018graph} incorporated attention mechanisms and later HGT \citep{hu2020heterogeneous} extends GAT to heterogeneous graphs. These developments have established GNNs as a powerful tools for relational  modeling \citep{gao2023survey}.

\paragraph{Temporal graph.}  
As described in \citet{gastinger2024tgb}, temporal graph methods fall into two broad categories: discrete-time and continuous-time. Discrete-time methods exist for both homogeneous \citep{han2020graph} and heterogeneous datasets \citep{yin2019dhne,xue2021modeling,bian2019network}. Continuous-time methods arguably preserve more information and can be converted into discrete graphs, but the reverse is not possible \citep{kazemi2020representation}. TGN \citep{tgn_icml_grl2020} introduces a general framework for modeling continuous-time dynamic graphs, categorizing DyRep \citep{trivedi2019dyrep} as a special case, followed by several other models such as DyGFormer \cite{yu2023towards}, NAT \cite{luo2022neighborhood}, TNCN \cite{zhang2406efficient}, and CTAN \cite{gravina2024long}.
Other temporal graph models such as HTGN-BTW \citep{yue2022htgn} and STHN \citep{li2023simplifying} propose different extensions of TGN to handle heterogeneous data. Beyond the standard temporal graphs, several methods have also been proposed for modeling temporal knowledge graphs \citep{li2021temporal,jin2019recurrent,li2022complex}.

\subsection{Benchmark datasets.}

\label{sec:benchmark-dataset}

Several benchmark efforts have been proposed across related areas. The temporal graph benchmarks include the TGB \citep{huang2023temporal}, its heterogeneous and knowledge graph extension (TGB 2.0) \citep{gastinger2024tgb}, and TGB-Seq \citep{yi2025tgb}, which include a more complex sequence of edge dynamics. 
For static graph learning, OGB \citep{hu2020open} and its large-scale extension OGB-LSC \citep{hu2021ogb} provide widely used benchmarks. In the recommendation domain, large-scale user-item interaction datasets have been released through benchmarks such as MIND \citep{wu2020mind}, TenRec \citep{yuan2022tenrec}, NineRec \citep{zhang2024ninerec}, and BARS \citep{zhu2022bars}. For event sequence modeling and temporal point processes (TPP), recent benchmarks include EBES \citep{osin2024ebes}, EasyTPP \citep{xue2023easytpp}, and HOTPP \citep{karpukhin2024hotpp}.

\looseness=-1
The closest dataset or benchmark work from our paper is the temporal graph benchmark (TGB) by \citet{huang2023temporal}. It contains several datasets that model user behaviors. These datasets can be roughly divided into two categories: (1) user-to-user interaction datasets, such as \textsc{tgbl-coin}, \textsc{tgbl-comment}, and \textsc{tgbn-trade}; (2) user-to-item bipartite interaction datasets, such as \textsc{tgbl-wiki}, \textsc{tgbl-review}, \textsc{tgbn-genre}, \textsc{tgbn-reddit}, and \textsc{tgbn-token}. The first category of user-to-user interaction datasets are similar to the relational event part of our datasets; however our datasets also contain personal event sequences, in addition to relational events, which are not present in the TGB dataset.

The second category of TGB datasets are bipartite temporal graphs. In our PRES formulation, the interaction of a user to an item can be encoded as a personal event, where an item is represented as an event or a token. However, the personal event abstraction in the PRES formulation can also encode other types of events. An example is illustrated in Figure \ref{fig:pres}, where User A has both user-item events (product views) and other types of events such as ‘Add to cart’ and ‘Purchase’. Our formulation also contains relational events that model use-to-user interactions.

In addition, formulating an item as a personal event instead of a node enables the flexibility of encoding items that have hierarchical information such as Geohash in our Brightkite dataset. Each character in the geohash encodes increasingly detailed location information. When we encode an 8-letter geohash as a node, we lose the hierarchical information encoded in the geohash. In contrast, if we are not forced to represent an event as a node, we have more flexibility to encode the hierarchical structure of the geohash. For example, in a sequence model, one could tokenize the event freely. A single event could be encoded into multiple tokens.

In terms of the size, the TGB dataset ranges from a small size of 255 node graph (\textsc{tgbn-trade}) to nearly a million node graph (\textsc{tgbl-comment}). Our PRES dataset also ranges from a small-to-medium size of 58 thousand users (\textsc{pres-brightkite}) to a relatively large dataset of \textsc{pres-github} with 3.6 million users. In terms of the number of events (or number of edges in TGB dataset) our PRES datasets are comparable with TGB datasets, and in some cases larger than TGB datasets. The number of edges in TGB datasets range from around 150 thousands edges (\textsc{tgbl-wiki}) to 44 million edges (\textsc{tgbl-comment}); whereas the number of events in our datasets range from 1.5 million (\textsc{pres-amazon-clothing}) to more than 100 million events (\textsc{pres-github}).

\subsection{Other research on graph and sequence.} Several studies have been conducted on different settings on temporal and structural dynamics. Some models focus on modeling graph and time series data using spatio-temporal graph \citep{yu2017spatio,yu2020spatio,bai2020adaptive,wu2020connecting}. Other model combine graph model output and sequence model output in various application areas \citep{hai2024study,shi2019two,mao2021molecular}. 
A recent study focus on tokenizing graph and applying transformers or state space models (SSMs) for graph learning \citep{zhang2001graph,wang2024graph,behrouz2024graph,huang2024can, ding2024dygmamba,li2024dyg,behrouz2024best}.
Another studies works on incorporating knowledge graph into language model. 
\citep{liu2020k,yasunaga2021qa,huang2020knowledge}, as well as performing graph-augmented language generation \citep{peng2024graph,zhu2025graph}.



\section{Experiment Details}
\label{sec:ap-experiment-details}

\subsection{Hyperparameters}

\paragraph{Personal event prediction task.}
In Table~\ref{tab:hyperparams_pers_events}, we present the hyperparameters used during the training of various models for personal event prediction tasks. We use the following notations: \textbf{Emb\ Dim} denotes the dimensionality of token embeddings; \textbf{Heads} is the number of attention heads; \textbf{Layers} refers to the number of hidden layers; \textbf{Channels} indicates the number of hidden channels per layer in GAT and GCN models; \textbf{Max\ Examples} is the maximum number of training samples generated per user; \textbf{Num\ Neg\ Samples} represents the number of negative samples for each (positive) sample; and \textbf{Num\ Neighbors} is the number of neighbors sampled per layer for GNN models. Additionally, due to GPU memory limitations, we reduce the embedding dimensions for the TGN and DyRep models to 16 and 32, respectively, for the \textsc{pres-brightkite} and \textsc{pres-gowalla} datasets, and to 32 and 64 for \textsc{pres-amazon-clothing} and \textsc{pres-amazon-electronics}.

\begin{table}[t]
  \centering
  \scriptsize
  \caption{Hyperparameter configurations for personal event prediction tasks}
  \setlength{\tabcolsep}{3pt}
  \resizebox{\textwidth}{!}{%
    \begin{tabular}{lccccccccccc}
\toprule
\textbf{Model Name} &
\shortstack{\textbf{Learning} \\ \textbf{Rate}} &
\shortstack{\textbf{Batch} \\ \textbf{Size}} &
\textbf{Epochs} &
\shortstack{\textbf{Emb} \\ \textbf{Dim}} &
\textbf{Heads} &
\textbf{Layers} &
\textbf{Channels} &
\shortstack{\textbf{Max} \\ \textbf{Events}} &
\shortstack{\textbf{Max} \\ \textbf{Examples}} &
\shortstack{\textbf{Num Neg} \\ \textbf{Samples}} &
\shortstack{\textbf{Num} \\ \textbf{Neighbors}} \\
\midrule
      BERT & 3e-4 & 1024 & 10 & 64 & 4 & 4 & -- & 100 & 50 & 10 & -- \\
      BERT-n2v-p & 3e-4 & 1024 & 10 & 64 & 4 & 4 & -- & 100 & 50 & 10 & -- \\
      BERT-n2v-i & 3e-4 & 1024 & 10 & 64 & 4 & 4 & -- & 100 & 50 & 10 & -- \\      
      \midrule      
      GCN & 1e-3 & 1024 & 10 & 128 & -- & 2 & 128 & 100 & -- & 5 & 10 \\
      GCN\_PR & 1e-3 & 1024 & 10 & 128 & -- & 2 & 128 & 100 & -- & 5 & 10  \\
      \midrule      
      GAT & 1e-3 & 1024 & 10 & 128 & 2 & 2 & 64 &  100 & -- & 5 & 10\\
      GAT\_PR & 1e-3 & 1024 & 10 & 128 & 2 & 2 & 64 & 100 & -- & 5 & 10 \\
      \midrule      
      TGN & 1e-3 & 4096 & 10 & 16/32 & -- & -- & -- & 100 & -- & 5 & 10\\
      TGN\_PR & 1e-3 & 4096 & 10 & 16/32 & -- & -- & -- & 100 & -- & 5 & 10\\
      \midrule      
      DyRep & 1e-3 & 4096 & 10 & 32/64 & -- & -- & -- & 100 & -- & 5 & 10\\
      DyRep\_PR & 1e-3 & 4096 & 10 & 32/64 & -- & -- & -- & 100 & -- & 5 & 10\\
      \bottomrule
    \end{tabular}
  } 
  \label{tab:hyperparams_pers_events}
\end{table}

\paragraph{Relational event prediction task.}
In Table~\ref{tab:hyperparams_rel_events}, we present the hyperparameters used across all models for relational event prediction tasks. Due to memory and time constraints, batch size, number of epochs, and embedding dimensions were adjusted per dataset. All datasets used a batch size of 4096, except for \textsc{pres-github}, which used 512. The number of training epochs was set to 5 for \textsc{pres-github}, 20 for \textsc{pres-gowalla} and \textsc{pres-amazon-electronics}, 100 for \textsc{pres-amazon-clothing}, and 1000 for \textsc{pres-brightkite}. The model checkpoint with the best validation MRR was saved and used for testing. As shown in our results, TGN, DyRep and TNCN could not be run on \textsc{pres-github}. For the remaining datasets, the embedding dimension for TGN and DyRep was 128, except for \textsc{pres-gowalla}, which used 64 to avoid GPU out-of-memory errors. TNCN used `NCN\_mode` of 1, an embedding dimension of 64 and a smaller batch size (1024) for all datasets.

\begin{table}[t]
  \centering
  \scriptsize
  \caption{Hyperparameter configurations for relational event prediction tasks}
  \resizebox{\textwidth}{!}{%
    \begin{tabular}{lccccccccc}
\toprule
\textbf{Model Name} &
\shortstack{\textbf{Learning} \\ \textbf{Rate}} &
\shortstack{\textbf{Batch} \\ \textbf{Size}} &
\textbf{Epochs} &
\shortstack{\textbf{Emb} \\ \textbf{Dim}} &
\textbf{Heads} &
\textbf{Layers} &
\textbf{Channels} &
\shortstack{\textbf{Num Neg} \\ \textbf{Samples}} &
\shortstack{\textbf{Num} \\ \textbf{Neighbors}} \\
\midrule

      GCN & 1e-3 & 512/4096 & 5-1000 & 128 & -- & 2 & 128 & 5 & 10 \\
      GCN\_PR & 1e-3 & 512/4096 & 5-1000 & 128 & -- & 2 & 128 & 5 & 10  \\
      GCN+S & 1e-3 & 512/4096 & 5-1000 & 128 & -- & 2 & 128 & 5 & 10  \\
      \midrule
      GAT & 1e-3 & 512/4096 & 5-1000 & 128 & 2 & 2 & 128 & 5 & 10 \\
      GAT\_PR & 1e-3 & 512/4096 & 5-1000 & 128 & 2 & 2 & 128 & 5 & 10 \\
      GAT+S & 1e-3 & 512/4096 & 5-1000 & 128 & 2 & 2 & 128 & 5 & 10 \\      
      \midrule
      TConv & 1e-3 & 512/4096 & 5-1000 & 128 & 2 & 2 & 128 & 5 & 10 \\
      TConv\_PR & 1e-3 & 512/4096 & 5-1000 & 128 & 2 & 2 & 128 & 5 & 10 \\
      TConv+S & 1e-3 & 512/4096 & 5-1000 & 128 & 2 & 2 & 128 & 5 & 10 \\      
      \midrule
      TGN & 1e-3 & 4096 & 20-1000 & 64/128 & -- & -- & 128 & 5 & 10 \\      
      DyRep & 1e-3 & 4096 & 20-1000 & 64/128 & -- & -- & 128 & 5 & 10 \\      
      TNCN & 1e-3 & 1024 & 20-1000 & 64 & -- & -- & 128 & 5 & 10 \\
      \bottomrule
    \end{tabular}
  } 
  \label{tab:hyperparams_rel_events}
\end{table}

\begin{table}[t]
\centering
\scriptsize
\caption{Computational Time (in hours) for Different Models and Datasets}
\label{tab:exp_times}
\begin{tabular}{l|ccccc}
\toprule
\multirow{2}{*}{Method} & \multicolumn{5}{c}{Time (h)} \\
\cmidrule{2-6}
 & \textsc{amazon-clothing} & \textsc{amazon-electronics} & \textsc{brightkite}  & \textsc{gowalla}  &  \textsc{github} \\
\midrule
\multicolumn{6}{l}{\textcolor{gray}{Relational event prediction tasks}} \\
\midrule
GCN & 0.06{\scriptsize $\pm$0.00} & 0.05{\scriptsize $\pm$0.00} & 0.60{\scriptsize $\pm$0.00} & 0.26{\scriptsize $\pm$0.00} & 8.38{\scriptsize $\pm$0.11} \\
GCN\_RP & 0.10{\scriptsize $\pm$0.01} & 0.17{\scriptsize $\pm$0.00} & 0.40{\scriptsize $\pm$0.00} & 1.98{\scriptsize $\pm$0.03} & 7.39{\scriptsize $\pm$0.06} \\
GCN+S & 0.07{\scriptsize $\pm$0.00} & 0.05{\scriptsize $\pm$0.00} & 0.61{\scriptsize $\pm$0.00} & 0.29{\scriptsize $\pm$0.00} & 8.58{\scriptsize $\pm$0.12} \\
GAT & 0.07{\scriptsize $\pm$0.01} & 0.08{\scriptsize $\pm$0.02} & 0.61{\scriptsize $\pm$0.00} & 0.29{\scriptsize $\pm$0.00} & 8.41{\scriptsize $\pm$0.12} \\
GAT\_RP & 0.15{\scriptsize $\pm$0.03} & 0.18{\scriptsize $\pm$0.01} & 0.49{\scriptsize $\pm$0.06} & 2.07{\scriptsize $\pm$0.02} & 7.40{\scriptsize $\pm$0.06} \\
GAT+S & 0.09{\scriptsize $\pm$0.02} & 0.05{\scriptsize $\pm$0.00} & 0.62{\scriptsize $\pm$0.00} & 0.32{\scriptsize $\pm$0.00} & 2.52{\scriptsize $\pm$0.20} \\
TConv & 0.05{\scriptsize $\pm$0.00} & 0.04{\scriptsize $\pm$0.00} & 0.31{\scriptsize $\pm$0.00} & 0.50{\scriptsize $\pm$0.00} & 19.8{\scriptsize $\pm$5.38} \\
TConv\_PR & 0.10{\scriptsize $\pm$0.00} & 0.21{\scriptsize $\pm$0.00} & 1.86{\scriptsize $\pm$0.00} & 2.98{\scriptsize $\pm$0.04} & 12.5{\scriptsize $\pm$6.83} \\
TConv+S & 0.05{\scriptsize $\pm$0.00} & 0.04{\scriptsize $\pm$0.00} & 0.33{\scriptsize $\pm$0.00} & 0.57{\scriptsize $\pm$0.00} & 20.2{\scriptsize $\pm$8.79} \\
TGN & 0.49{\scriptsize $\pm$0.03} & 0.32{\scriptsize $\pm$0.00} & 1.06{\scriptsize $\pm$0.01} & 4.62{\scriptsize $\pm$0.10} & -- \\
DyRep & 0.49{\scriptsize $\pm$0.02} & 0.31{\scriptsize $\pm$0.00} & 1.03{\scriptsize $\pm$0.01} & 4.43{\scriptsize $\pm$0.07} & -- \\
TNCN & 0.63{\scriptsize $\pm$0.01} & 0.52{\scriptsize $\pm$0.00} & 1.37{\scriptsize $\pm$0.02} & 2.04{\scriptsize $\pm$0.05} & -- \\
\midrule
\multicolumn{6}{l}{\textcolor{gray}{Personal event prediction tasks}} \\
\midrule
GCN & 5.81{\scriptsize $\pm$0.10} & 7.14{\scriptsize $\pm$0.29} & 1.73{\scriptsize $\pm$0.02} & 8.01{\scriptsize $\pm$0.71} & -- \\
GCN\_PR & 5.80{\scriptsize $\pm$0.11} & 7.21{\scriptsize $\pm$0.29} & 1.73{\scriptsize $\pm$0.03} & 7.45{\scriptsize $\pm$1.82} & -- \\
GAT & 5.83{\scriptsize $\pm$0.10} & 7.17{\scriptsize $\pm$0.28} & 1.73{\scriptsize $\pm$0.03} & 7.33{\scriptsize $\pm$1.82} & -- \\
GAT\_PR & 5.82{\scriptsize $\pm$0.10} & 7.24{\scriptsize $\pm$0.30} & 1.76{\scriptsize $\pm$0.02} & 7.51{\scriptsize $\pm$1.79} & -- \\
TGN & 3.94{\scriptsize $\pm$0.40} & 4.10{\scriptsize $\pm$0.10} & 0.33{\scriptsize $\pm$0.01} & 3.12{\scriptsize $\pm$0.75} & -- \\
TGN\_PR & 4.38{\scriptsize $\pm$0.39} & 5.75{\scriptsize $\pm$0.19} & 0.81{\scriptsize $\pm$0.03} & 7.89{\scriptsize $\pm$1.78} & -- \\
DyRep & 2.03{\scriptsize $\pm$0.38} & 3.23{\scriptsize $\pm$0.34} & 0.38{\scriptsize $\pm$0.01} & 4.11{\scriptsize $\pm$1.03} & -- \\
DyRep\_PR & 4.88{\scriptsize $\pm$0.10} & 5.96{\scriptsize $\pm$0.20} & 0.78{\scriptsize $\pm$0.03} & 7.85{\scriptsize $\pm$1.86} & -- \\
BERT & 4.67{\scriptsize $\pm$0.06} & 6.30{\scriptsize $\pm$0.15} & 2.65{\scriptsize $\pm$0.02} & 9.21{\scriptsize $\pm$1.11} & -- \\
BERT+n2v-i & 3.41{\scriptsize $\pm$0.14} & 4.43{\scriptsize $\pm$0.19} & 2.54{\scriptsize $\pm$0.01} & 6.40{\scriptsize $\pm$0.19} & -- \\
BERT+n2v-p & 3.60{\scriptsize $\pm$0.22} & 4.78{\scriptsize $\pm$0.14} & 2.52{\scriptsize $\pm$0.01} & 6.38{\scriptsize $\pm$0.20} & -- \\
\bottomrule
\end{tabular}%
\end{table}


\subsection{Computing Resources}
We conducted all experiments on a server equipped with 8 NVIDIA Ampere A10G GPUs (24 GB each), 16 CPU cores, and a RAM upper limit of 512 GB. To fully leverage all resources, we parallelized the training runs so that each experiment used a single GPU. Each experiment is designed to be run on a single-GPU machine. Table~\ref{tab:exp_times} summarizes the average training time (in hours) and standard deviation for each model across five datasets, categorized by task type. For relational event prediction tasks, lightweight GCN and GAT variants exhibit minimal computational overhead, with training times generally under one hour except on the GitHub dataset. In contrast, temporally expressive models such as TGN and DyRep incur significantly higher costs, especially on large-scale datasets like Gowalla. In personal event prediction tasks, training times increase across the board, with most models exceeding 7 hours on larger datasets, again highlighting the computational demands of modeling fine-grained temporal dynamics.

\begin{figure}[h]
    \centering
    \includegraphics[width=\textwidth]{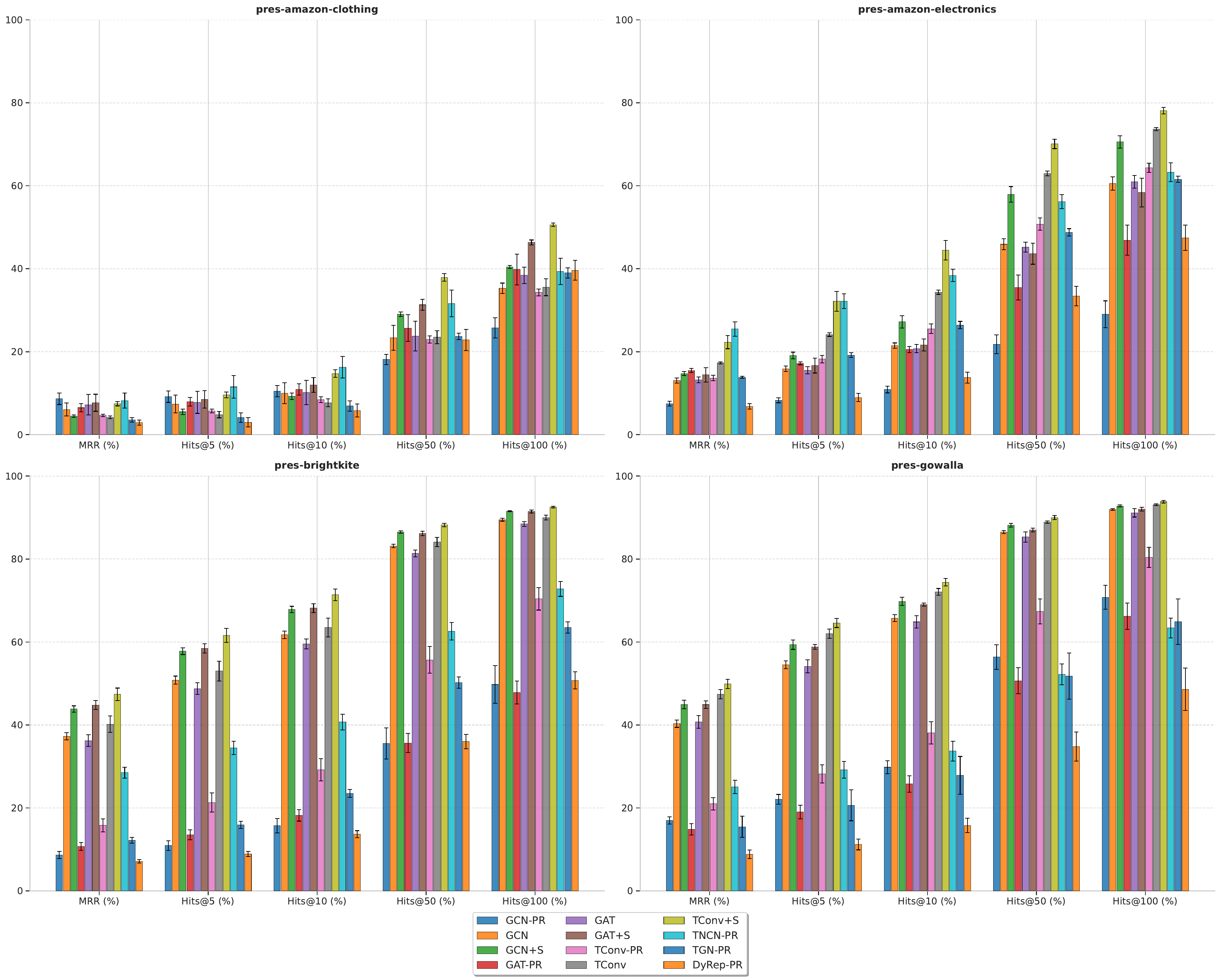}
    \caption{Comparison of relational event predictions across different datasets.}
    \label{fig:relational_all}
\end{figure}


\section{Additional Experimental Results}
\label{sec:ap-experiments}

\begin{figure}[t]
    \centering
    \includegraphics[width=\textwidth]{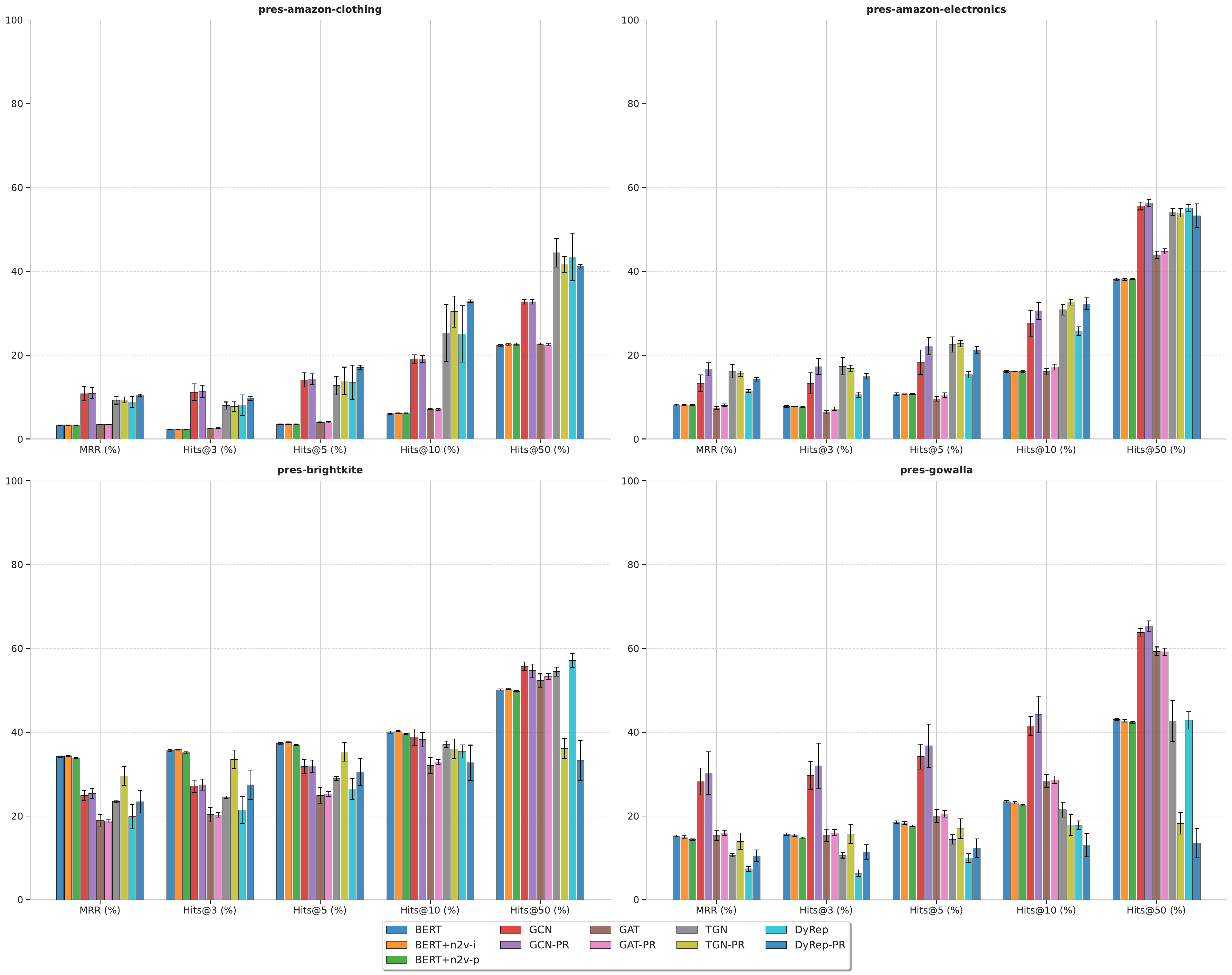}
    \caption{Comparison of personal event predictions across different datasets.}
    \label{fig:personal_all}
\end{figure}

\paragraph{Experiment Figures}
In Figures~\ref{fig:relational_all} and~\ref{fig:personal_all}, we present the results from the main paper in a more visual format to facilitate comparison across methods. In the relational event prediction tasks, across all datasets and metrics, static GNNs augmented with personal event sequence embeddings (GCN+S, GAT+S, and especially TConv+S) consistently perform well, achieving the best or second-best results. This highlights the benefit of integrating both personal and relational signals. For temporal graph methods, the TGN and DyRep under-perform in most of datasets and most metrics. TNCN perform well on \texttt{amazon} datasets on MRR and Hits@k with lower $k$, but under-perform on other metrics or other datasets. 
For personal event prediction tasks, BERT+n2v-i offers slight improvements over regular BERT. In particular, BERT-based models exhibit competitive performance in some cases, most prominently on the Brightkite dataset, where they outperform GNN-based counterparts at MMR and lower hit rate thresholds such as Hits@3, Hits@5, and Hits@10.

\paragraph{Additional analysis}


One of the main takeaways of the paper is that models leveraging both personal and relational events outperform those using only one type in either relational or personal event prediction tasks. For example, in relational event prediction, the "+S" models (GCN+S, GAT+S, and TConv+S) incorporate sequence embeddings of personal event data into the relational event graph, boosting performance over models that rely solely on relational events (GCN, GAT, or TConv). This highlights the need for models that jointly account for both signals. 

We then compare the "+S" strategy to the "\_RP" models, which convert each unique personal event into a node and add it to the user-to-user graph used by the GCN model, creating edges between users and their personal event nodes. In most cases, "+S" models outperform "\_RP" models, with very few exceptions.  
In some datasets, such as \texttt{pres-brightkite}, \texttt{pres-gowalla}, and \texttt{pres-github}, adding personal event nodes to the graph even reduces performance. These results may be explained as follows:
\begin{itemize}
    \item When we exclude personal events (standard GCN, GAT, and TConv), the model is still able to extract and learn some predictive information from the user-to-user relational events alone to some degree.
    \item When we include personal events as nodes (GCN\_RP, GAT\_RP, TConv\_RP), this adds more noise than benefit to the system, as the number of personal event nodes is much larger than the number of user nodes. As a result, model performance decreases.
    \item When we encode personal events as sequence embeddings (GCN+S, GAT+S, and TConv+S), this produces meaningful features without introducing excessive noise. The models are able to capture additional signals from these personal event embeddings, leading to performance improvements.
    \item In addition, when modeling personal events using a sequence model (BERT) in the "+S" strategy, we retain the hierarchical information of the personal events (such as geohash check-in events in \texttt{pres-brightkite} and \texttt{pres-gowalla}). In contrast, when we convert personal events into nodes as in the "\_RP" strategy, we lose the hierarchical information present in the events.
\end{itemize}


However, this pattern does not always generalize to every dataset, as we see in \texttt{pres-amazon-clothing}, where "\_RP" models perform relatively well on MRR and Hits@5, but not on Hits@k metrics with larger $k$. This suggests that in this dataset, personal event nodes may not merely act as noise in the graphs. Instead, they help the model improve precision on the top candidates (i.e., fewer but more accurate suggestions), at the expense of lower recall coverage for larger $k$. In addition, the encoded hierarchical information in the product-rating nodes may be less important in this dataset.  
This observation may influence the architecture design of future models aiming to leverage both personal and relational event signals.

\section{Broader Impacts}

\label{sec:impact}


\paragraph{Broader impact}

The datasets and prediction tasks we release may support future research on user event modeling, particularly in settings that involve both personal and relational events. Researchers can build models on top of these resources and evaluate them in a consistent way. This can help accelerate empirical progress and facilitate more comparable results. This has potential impact in a range of industry applications where modeling user behavior is critical, such as recommendation, fraud detection, and user interaction analysis.

\paragraph{Potential negative impact}
The datasets we release may not cover all use cases of user event modeling, and may reflect only a subset of real-world scenarios. This could introduce bias in model development or evaluation, especially if models are tuned specifically for the structure or properties of our datasets. As a result, there is a risk that future methods may overfit to our datasets and generalize less effectively to other domains or applications.



\end{document}